\documentclass{article}
\usepackage{iclr2027_conference,times}

\usepackage{amsmath,amsfonts,bm}

\def\eqref#1{equation~\ref{#1}}

\def\1{\bm{1}}

\DeclareMathAlphabet{\mathsfit}{\encodingdefault}{\sfdefault}{m}{sl}
\SetMathAlphabet{\mathsfit}{bold}{\encodingdefault}{\sfdefault}{bx}{n}

\newcommand{\E}{\mathbb{E}}

\newcommand*{\hole}{{\mspace{1mu}\cdot\mspace{1mu}}}

\usepackage{amssymb}
\usepackage{bm}
\usepackage{mathtools}
\usepackage{makecell}
\usepackage{array}
\usepackage{booktabs}
\usepackage{multirow}
\usepackage{graphicx}
\usepackage{hyperref}
\usepackage{tabularx}
\usepackage{siunitx}
\usepackage{subcaption}
\usepackage{collcell} 
\usepackage{url}
\usepackage[capitalise,noabbrev]{cleveref}
\usepackage{algorithm}
\usepackage{algpseudocode}
\usepackage{wrapfig}
\usepackage[table]{xcolor}

\usepackage{svg}

\definecolor{bestcell}{RGB}{144,238,144}

\newcolumntype{M}[2]{%
  >{\def\meanformat{#1}\def\sdformat{#2}%
    \collectcell\formatMeanSD}c<{\endcollectcell}%
}
\NewDocumentCommand{\formatMeanSD}{>{\SplitArgument{1}{\pm}}m}{%
  \formatMeanSDparts #1%
}
\NewDocumentCommand{\formatMeanSDparts}{m m}{%
  \IfNoValueTF{#2}{#1}{
    \begingroup
      \sisetup{mode=text, reset-text-series=false}%
      \tablenum[
        table-format=\meanformat,
        table-model-setup=\bfseries
      ]{#1}%
      \mdseries\unboldmath\(\,\pm\,\)%
      \tablenum[
        table-format=\sdformat,
        table-model-setup=\mdseries 
      ]{#2}%
    \endgroup
  }%
}

\newcommand{\setlastcellformat}[1]{%
    \def\lastcellformat{#1}%
}
\newcommand{\cellcolorlast}[2]{%
    \expanded{%
        \noexpand\multicolumn{1}{%
            >{\noexpand\columncolor{#1}[\tabcolsep][0pt]}%
            \lastcellformat @{}%
        }{\unexpanded{#2}}%
    }%
}

\author{
    Meir Roketlishvili$^{1,*}$,
    Felix Huber$^{1,*}$,
    Wenda Chu$^{2}$, 
    \textbf{Yisong Yue}$^{2}$,
    \textbf{Raul Astudillo}$^{1}$ \\[0.5em]
    $^{1}$Mohamed bin Zayed University of Artificial Intelligence, Abu Dhabi, UAE \\
    $^{2}$California Institute of Technology, California, USA \\[0.5em]
    {\small\texttt{\{meir.roketlishvili,felix.huber,raul.astudillo\}@mbzuai.ac.ae}} \\
    {\small\texttt{\{wchu,yyue\}@caltech.edu}}
}

\title{Multi-Objective Protein Optimization via\\ Modular Diffusion Steering}
\iclrfinalcopy

\begin{document}

\maketitle
\lhead{Preprint.}

\begingroup
\renewcommand{\thefootnote}{\fnsymbol{footnote}}
\footnotetext[1]{These authors contributed equally to this work.}
\endgroup

\begin{abstract}
Protein engineering often requires improving several properties simultaneously, motivating the search for high-performing sequences that represent useful fitness tradeoffs. Discovering these sequences is challenging because the design space is combinatorially large and experimental campaigns often permit only a few rounds of batched measurements. We introduce \textit{MODS}, a modular framework for multi-objective protein optimization via inference-time diffusion steering. For each proposal, MODS samples augmented Chebyshev scalarization weights and a fitness-surrogate ensemble member to define a scalar steering reward. Varying the weights directs generation toward different objective tradeoffs, while varying ensemble members diversifies the predictions used for guidance. Because MODS expresses multi-objective acquisition through scalar rewards, it can be paired with different diffusion models and compatible steering procedures. After each evaluation batch, MODS updates the surrogates without retraining the generative model. We evaluate MODS on six protein optimization tasks using predictive oracles trained on experimental assay data, with two diffusion models and two steering procedures.
MODS achieves competitive final hypervolume across all six tasks with both diffusion models, and substantially outperforms the best evaluated baseline on several tasks.
Our results demonstrate a flexible and effective approach to multi-objective protein optimization with limited feedback.
\end{abstract}

\section{Introduction}
\label{sec:introduction}

\begin{figure}[b]
    \centering
    \includegraphics[width=0.77\linewidth]
        {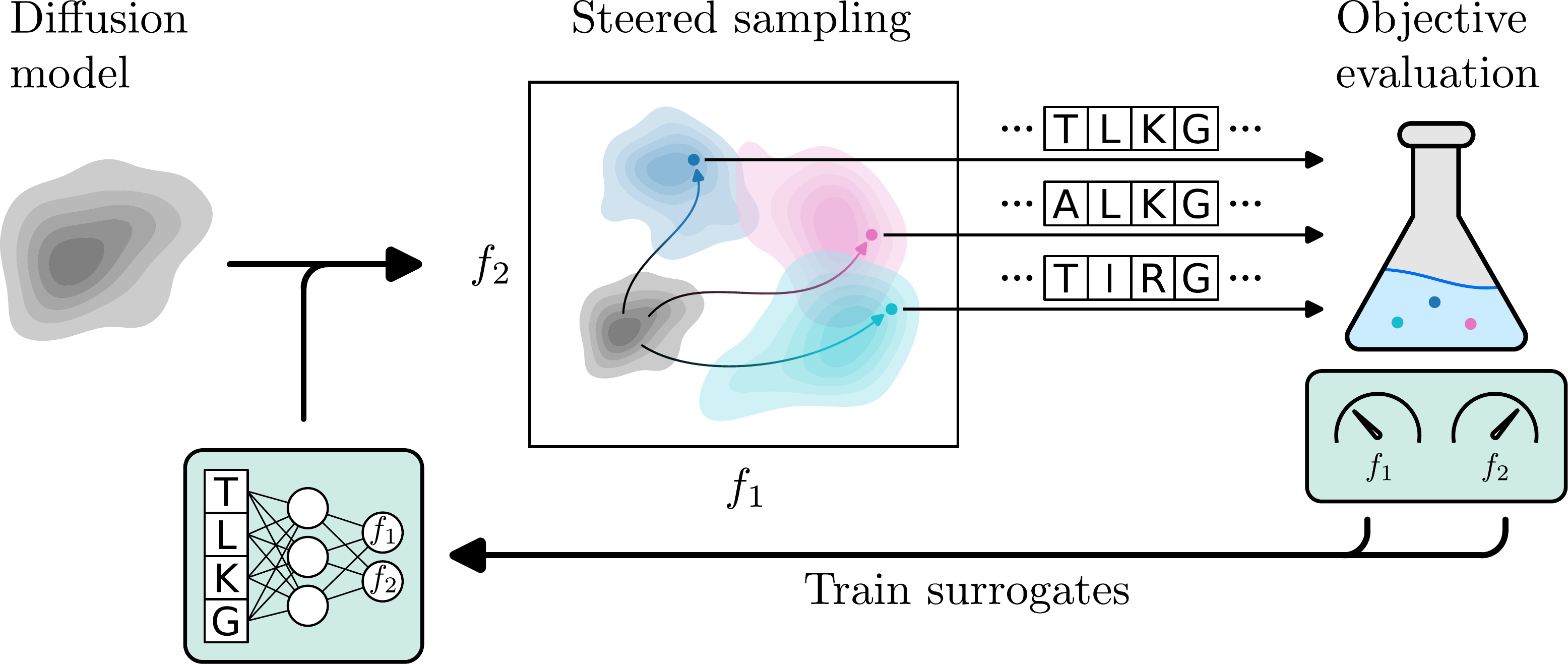}
    \caption{
        \textbf{Overview of MODS.}
        At each optimization round, MODS samples Chebyshev scalarization weights and a surrogate-ensemble member to construct rewards targeting different objective tradeoffs under predictive uncertainty.
        These rewards steer a fixed diffusion model to generate the next evaluation batch.
        The generated sequences are evaluated, and their fitness observations are used to update the surrogate ensemble for the following round.
    }
    \label{fig:illustrate_method}
\end{figure}

Protein engineering often requires improving several properties simultaneously. Because improving one property can degrade another, the goal is not necessarily a single best sequence but a set of high-performing sequences representing useful fitness tradeoffs~\citep{Gruver2023Protein,Luo2025MosPro}. Multi-objective optimization formalizes this goal through the Pareto front, which represents solutions for which no objective can be improved without worsening at least one other. Finding these tradeoffs is difficult because the protein sequence space grows exponentially with sequence length, while experimental costs severely limit the number of variants that can be evaluated~\citep{Yang2019MachineLearning}. Protein engineering campaigns therefore often permit only a few rounds of batched measurements, and each batch must be selected with incomplete knowledge of the fitness landscapes.

Generative protein models make this search more tractable by learning patterns from unlabeled sequences~\citep{Riesselman2018DeepGenerative,alamdari2023evodiff}. The resulting generative models focus generation on plausible regions of sequence space but do not encode the properties required for a particular engineering task. For single-objective optimization, prior work has used task-specific fitness predictions to steer fixed generative models, refine samples at inference time, or adapt generators toward high-reward sequences~\citep{Yang2025Steering,Li2025SVDD,Uehara2025Refinement,Wang2025DRAKES}. These approaches combine sequence knowledge with task-specific fitness signals.

Multi-objective generative optimization must produce candidates spanning a range of Pareto tradeoffs rather than concentrate generation on a single objective or fixed preference. Prior work has addressed this challenge by combining learned sequence representations with Bayesian optimization, guiding protein generation through hypervolume-based acquisition or gradients from multiple property predictors, and learning generative distributions over different objective preferences~\citep{Stanton2022Accelerating,Gruver2023Protein,Luo2025MosPro,Jain2023MultiObjectiveGFlowNets,Yuan2025ParetoFlow,Annadani2025PreferenceGuided}. These methods provide effective solutions but generally instantiate the acquisition strategy together with a particular generative model or optimization procedure. This motivates the need for a principled, modular approach to multi-objective generative optimization that can be applied to any compatible combination of generative model and steering mechanism.

We introduce \textit{MODS}, an iterative, modular approach to multi-objective protein optimization through inference-time diffusion steering. MODS combines randomized augmented Chebyshev scalarizations with surrogate ensemble sampling. Scalarization decomposes the multi-objective problem into a family of scalar steering problems targeting different objective tradeoffs~\citep{Knowles2006ParEGO,Paria2020Flexible}. For each proposal, MODS samples both scalarization weights and a fitness-surrogate ensemble member. The weights determine the targeted tradeoff, while the ensemble member selects a plausible predicted fitness landscape. Varying both across proposals couples exploration over objective tradeoffs with exploration under predictive uncertainty. Because each pair defines a scalar reward, MODS separates multi-objective acquisition from the implementation of generation, allowing it to be paired with different diffusion models and compatible steering procedures.

Figure~\ref{fig:illustrate_method} illustrates the resulting optimization loop. Each campaign begins with an initial set of evaluated sequences. At every subsequent round, MODS trains a surrogate ensemble on all accumulated observations, constructs rewards from sampled scalarization weights and ensemble members, and uses those rewards to steer the fixed generative model. The generated sequences are evaluated to expand the dataset for the next round. Thus, new feedback progressively refines the task-specific steering without retraining the generative model.

We evaluate MODS on six protein optimization tasks, five with two objectives and one with three, using fixed predictive oracles trained on experimental assay data. Each simulated campaign begins with 100 evaluated sequences and acquires eight additional batches of 100. We instantiate MODS with two diffusion models and two steering procedures, compare it with alternative optimization methods, and examine its sensitivity to the diffusion model, steering procedure, and steering strength. On every task, at least one MODS configuration substantially outperforms all evaluated baselines in final hypervolume, while the other configurations remain competitive.

Our main contributions are as follows:
\begin{itemize}
    \item \textbf{A modular framework for multi-objective generative optimization:} MODS represents acquisition under competing objectives and uncertain feedback as a mixture of reward-tilted sequence distributions, defined by scalarization weights and surrogate-ensemble members. Because each component is specified through a scalar reward, the framework can combine different pretrained diffusion models with compatible steering procedures.
    \item \textbf{A systematic evaluation under limited feedback:} Across six assay-derived oracle benchmarks, MODS consistently delivers substantial gains in final hypervolume over the evaluated baselines. Experiments with two diffusion models and two steering procedures further characterize sensitivity to the diffusion model, steering procedure, and steering strength.
\end{itemize}
\section{Related work}
\label{sec:related}

\paragraph{Protein generative models and steering.}
Deep generative models learn patterns of variation from protein sequences and thereby provide priors that concentrate generation on plausible regions of sequence space~\citep{Riesselman2018DeepGenerative,alamdari2023evodiff}. These generative models can be directed toward task-specific properties using gradients, value estimates, or iterative refinement during sampling~\citep{Uehara2025InferenceTime,Li2025SVDD,Uehara2025Refinement}. Related approaches instead adapt the generative model itself by optimizing it with respect to a learned reward~\citep{Wang2025DRAKES}. MODS adopts the inference-time paradigm: it updates task-specific fitness surrogates as feedback accumulates while keeping the generative model fixed.

\paragraph{Protein optimization with limited feedback.}
Feedback-efficient protein optimization commonly alternates between surrogate-guided batch selection and experimental measurements or prioritizes variants near known functional sequences~\citep{Yang2025ActiveLearning,Ren2022Proximal}. These approaches typically operate in localized mutation spaces, considering either a small set of mutable positions or variants that are separated from known sequences by relatively few mutations. Generative protein models help extend optimization to larger combinatorial spaces by using learned models to focus search on plausible proteins while permitting coordinated changes across many positions. Under limited feedback, prior work uses surrogate ensembles to steer a diffusion model toward high-fitness sequences for single-objective design~\citep{Yang2025Steering}. MODS builds on this prior work while addressing the additional challenges posed by multiple, potentially competing objectives.

\paragraph{Multi-objective acquisition strategies.}
Multi-objective acquisition must promote high objective values while covering useful tradeoffs. Hypervolume-based methods address both goals by evaluating the expected increase in dominated hypervolume~\citep{Daulton2020Differentiable}. Scalarization takes a different approach, converting the multi-objective problem into a family of scalar optimization problems. Randomized scalarizations have been studied extensively in Bayesian optimization, including in combination with uncertainty-aware acquisition~\citep{Knowles2006ParEGO,Paria2020Flexible}. Chebyshev scalarizations are particularly useful because they can recover tradeoffs on nonconvex Pareto fronts that weighted sums cannot~\citep{Miettinen1998Nonlinear,Lin2024Smooth}. In MODS, the scalarization weights select an objective tradeoff, while the sampled surrogate member selects a predicted fitness landscape. Because both choices enter only through the scalar reward, they can vary between proposals and optimization rounds while the generative model remains fixed.

\paragraph{Multi-objective generative protein optimization.}
Multi-objective protein-design methods have combined learned sequence representations with Bayesian optimization, steered discrete diffusion with hypervolume-based acquisition, and composed gradients from fixed property predictors to guide mutation sampling toward Pareto-optimal sequences~\citep{Stanton2022Accelerating,Gruver2023Protein,Luo2025MosPro}. Other studies investigate multi-property steering, tree-guided peptide generation, and iterative adaptation of protein generators~\citep{Xiong2026ProteinGuide,Tang2025PepTune,Lim2025SAGEProt}. More generally, reward-conditioned generative models can produce solutions spanning different objective tradeoffs, including in sequential evaluation settings~\citep{Jain2023MultiObjectiveGFlowNets,Steinberg2025AGPS}. Guided flows and diffusion models have also been developed for offline multi-objective optimization with fixed datasets~\citep{Yuan2025ParetoFlow,Annadani2025PreferenceGuided,Shrestha2026ParetoConditioned}. Like prior online approaches, MODS performs adaptive acquisition under limited feedback, updating its surrogate ensemble after each evaluation batch. A key contribution of MODS is its modular design, which allows its multi-objective acquisition strategy to be paired with different diffusion models and compatible steering mechanisms.








\section{Problem setup}
\label{sec:problem_setup}

We consider a task-specific protein design problem over sequences of fixed length $L$ and an amino acid alphabet $\mathcal A$. The sequence space is $\mathcal X=\mathcal A^L$. Each sequence $x\in\mathcal X$ has $M\geq2$ unknown fitness objectives, summarized by
\[
    \mathbf{f}(x)
    =
    \bigl(f_1(x), \ldots, f_M(x)\bigr)
    \in \mathbb{R}^{M},
\]
and all objectives are to be maximized. Because fitness evaluations are expensive, the optimizer can query only a limited number of sequences in a small number of batched rounds.

Because the objectives may conflict, no single sequence needs to maximize all of them. We therefore seek Pareto-optimal tradeoffs. A sequence $x$ dominates $x'$, written $x\succ_{\mathbf f}x'$, if $f_m(x)\geq f_m(x')$ for every objective $m$, with strict inequality for at least one. The Pareto set consists of the sequences that are not dominated by any other sequence:
\[
    \mathcal{P}^{\star}
    =
    \left\{
        x \in \mathcal{X}
        \;\middle|\;
        \nexists\, x' \in \mathcal{X} : x' \succ_{\mathbf{f}} x
    \right\}.
\]
The corresponding Pareto front is the image of this set in objective space,
$\mathcal F^{\star}=\{\mathbf f(x)\mid x\in\mathcal P^{\star}\}$.

To guide the search beyond the available fitness observations, we assume access to a pretrained diffusion model $p_0$, trained independently on unlabeled protein sequences. The diffusion model provides a distribution over sequence space and remains fixed throughout optimization.

Optimization begins with an initial evaluated dataset $\mathcal D_0$. At each round $t=1,\ldots,T$, the optimizer uses the diffusion model and accumulated observations in $\mathcal D_{t-1}$ to generate a batch $\mathcal B_t$ of $B$ sequences. Evaluating a sequence $x\in\mathcal B_t$ returns a possibly noisy observation $\mathbf y_x$ of its fitness vector $\mathbf f(x)$. After evaluating the batch, the dataset becomes
\[
    \mathcal{D}_t
    =
    \mathcal{D}_{t-1}
    \cup
    \{(x,\mathbf{y}_x)\mid x\in\mathcal{B}_t\}.
\]
After $T$ rounds, the campaign has a total budget of $|\mathcal D_0|+TB$ sequence evaluations. The goal is to use the pretrained diffusion model and limited feedback to identify evaluated sequences whose fitness vectors approach $\mathcal F^{\star}$ while representing diverse objective tradeoffs.

\section{The MODS Framework}
\label{sec:algorithm}

\begin{wrapfigure}[23]{r}{0.39\textwidth}
    \centering
    \vspace{-2.9ex}
    \includegraphics[width=\linewidth]
        {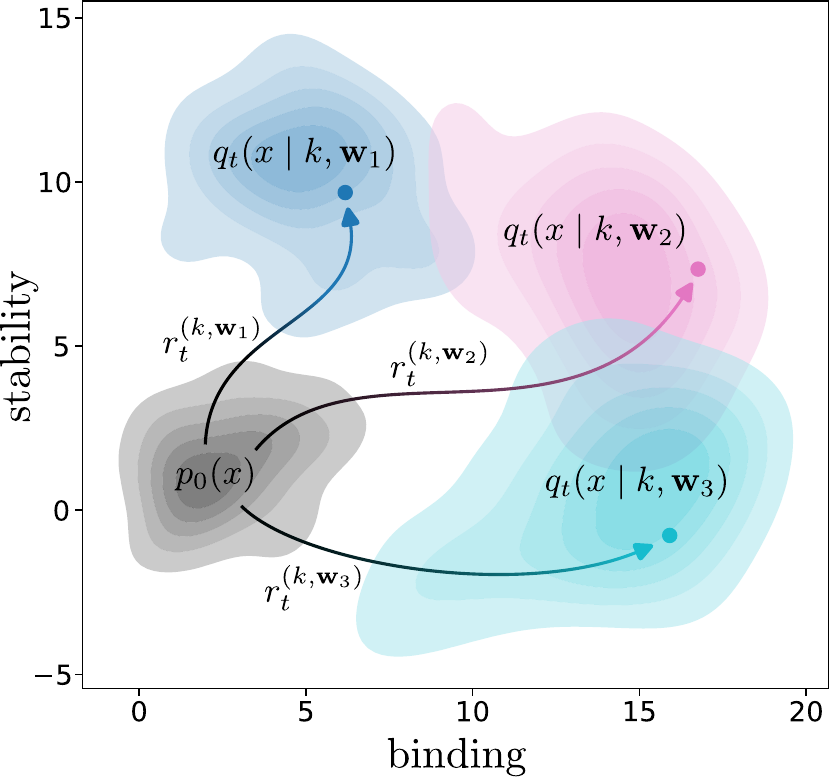}
    \caption{
        \textbf{Illustration of multi-objective steering in MODS.}
        Shown are the diffusion model \(p_0\) and three steered distributions from the sixth optimization round for the GB1 problem. To isolate the effect of scalarization weights, all three distributions use the same surrogate \(\widehat{\mathbf f}_6^{(k)}\) but different weight vectors \(\mathbf w_1\), \(\mathbf w_2\), and \(\mathbf w_3\), thereby targeting different regions of the objective space.
    }
    \label{fig:illustrate_steering}
\end{wrapfigure}

MODS is an iterative, modular framework for multi-objective protein optimization via inference-time diffusion steering. At each round, it fits an ensemble of fitness surrogates and applies randomized augmented Chebyshev scalarizations to sampled surrogate predictions, producing rewards that target different objective tradeoffs. Applied to the same fixed generative model, these rewards direct generation toward different regions of objective space, as illustrated in Figure~\ref{fig:illustrate_steering}. The resulting sequences form the evaluation batch, and their observed fitness values are used to update the surrogate ensemble for the next round.

\subsection{Surrogates and scalarized rewards}
\label{sec:scalarization}

Our goal is to generate sequences spanning different Pareto-optimal tradeoffs. The steering procedures considered here assign a scalar reward to each candidate, whereas the fitness surrogates predict a vector of objective values. Scalarization connects these two representations by converting the vector-valued predictions into a parametric family of per-sequence scalar rewards. Each weight vector specifies an objective tradeoff and thus defines a reward targeting a different region of the Pareto front. By varying the weights across candidates, MODS can pursue multiple tradeoffs in parallel.

At round $t$, we train an ensemble of $K$ fitness surrogates $\{\widehat{\mathbf f}_t^{(k)}\}_{k=1}^{K}$ on the observations in $\mathcal D_{t-1}$. Each ensemble member is a small multilayer perceptron that predicts all $M$ objectives, enabling repeated fitness prediction during generation without additional objective evaluations. Variation across ensemble predictions serves as a practical proxy for predictive uncertainty~\citep{Yang2025Steering}. Architecture and training details are provided in Appendix~\ref{sec:surrogate_details}.

Before scalarization, we normalize each predicted objective using the smallest and largest values of that objective observed in $\mathcal D_{t-1}$. For a normalized prediction vector $\mathbf{v} \in \mathbb{R}^M$ and a positive weight vector $\mathbf{w} \in \mathbb{R}^M$, we use the augmented Chebyshev scalarization
\begin{equation}
    s_{\mathbf{w}}(\mathbf{v})
    = \min_{m=1,\ldots,M} w_m v_m  +  \rho\sum_{m=1}^{M}w_m v_m,
    \label{eq:mods_scalarization}
\end{equation}
where $w_m>0$, $\sum_{m=1}^M w_m = 1$, and $\rho>0$ is a small augmentation coefficient. The minimum term emphasizes the smallest weighted objective, while the augmentation term also rewards improvements in the remaining objectives. Changing $\mathbf w$ changes the favored tradeoff. Unlike linear scalarizations, augmented Chebyshev scalarizations can recover optimal tradeoffs on nonconvex Pareto fronts~\citep{Miettinen1998Nonlinear}.

We sample a surrogate index $k \sim \operatorname{Uniform}(\{1,\ldots,K\})$ and scalarization weights $\mathbf{w} \sim \operatorname{Dirichlet}(\mathbf{1})$, and define the steering reward
\begin{equation}
    r_t^{(k,\mathbf w)}(x)
    = s_{\mathbf w}( \widehat{\mathbf f}_t^{(k)}(x) )
    \label{eq:mods_guidance}
\end{equation}
The weights specify the targeted objective tradeoff, while the sampled ensemble member selects one plausible predicted fitness landscape. Varying both across the evaluation batch promotes coverage of different tradeoffs and exploration under surrogate uncertainty.

This construction fits naturally with inference-time steering. For any fixed weight vector and surrogate prediction, the resulting scalar reward can be supplied to a compatible single-objective steering procedure without modifying or retraining the diffusion model. Changing the weight vector redirects generation toward a different tradeoff while reusing the same generative model and steering mechanism. Scalarization thus provides a modular interface between multi-objective acquisition and single-objective generative steering.

\subsection{Generating candidates with diffusion steering}
\label{sec:diffusion_steering}

For a fixed pair $(k,\mathbf w)$, MODS seeks a sampling distribution that attains high expected reward $r_t^{(k,\mathbf w)}$ while remaining close to the pretrained diffusion model. This tradeoff is expressed by the standard KL-regularized objective
\citep{Rafailov2023DPO}:
\begin{equation}
    q_t(\hole \mid k,\mathbf w)
    =
    \operatorname*{arg\,max}_{q\ll p_0}
    \Bigl(
        \E_{x\sim q}
        \bigl(r_t^{(k,\mathbf w)}(x)\bigr)
        -
        \alpha^{-1}\operatorname{KL}(q\,\|\,p_0)
    \Bigr).
    \label{eq:mods_regularized_objective}
\end{equation}
The optimizer has the analytic form
\begin{equation}
    q_t(x\mid k,\mathbf w)
    =
    \frac{
        p_0(x)\exp\bigl(\alpha r_t^{(k,\mathbf w)}(x)\bigr)
    }{Z_t^{(k,\mathbf w)}},
    \label{eq:mods_target}
\end{equation}
where $Z_t^{(k,\mathbf w)}$ is the normalizing constant. Larger $\alpha>0$ places greater emphasis on predicted reward, whereas smaller values keep the distribution closer to the diffusion model.

MODS uses inference-time diffusion steering~\citep{Uehara2025InferenceTime} to approximately sample from these reward-tilted distributions. Sampling different pairs $(k,\mathbf w)$ across proposals yields the mixture
$q_t(x)=\E_{k,\mathbf w}[q_t(x\mid k,\mathbf w)]$,
whose components target different objective tradeoffs under different surrogate predictions.

We instantiate this procedure with classifier guidance (CG)~\citep{dhariwal2021diffusion} and decoupled annealing posterior sampling (DAPS)~\citep{Zhang2025DAPS}. For CG, the surrogates predict fitness from corrupted sequences together with their noise levels. For DAPS, they are trained on clean sequences. For both steering procedures, the surrogates are trained only on fitness observations collected during the campaign rounds.

We generate the evaluation batch $\mathcal{B}_t$ in computational minibatches. Each minibatch shares a sampled surrogate member, while each candidate receives its own scalarization weights. This structure enables efficient batched surrogate evaluation. Both the surrogate member and scalarization weights remain fixed throughout each candidate's denoising trajectory.

\subsection{The complete MODS procedure}
\label{sec:iterative_bo}

We form the initial dataset $\mathcal D_0$ by sampling $n_0$ sequences from the diffusion model without steering and evaluating their fitness. Each subsequent round consists of three steps, as illustrated in Figure~\ref{fig:illustrate_method}:
\begin{enumerate}
    \item \textbf{Fit the surrogate ensemble.}
    Train fitness surrogates on all observations in
    $\mathcal D_{t-1}$, obtaining updated estimates of
    the objectives and their uncertainty.

    \item \textbf{Generate the evaluation batch.}
    Sample surrogate members and scalarization weights
    to construct steering rewards $r_t^{(k,\mathbf{w})}$.
    Use these rewards to steer the diffusion model and generate an evaluation
    batch $\mathcal{B}_t$ of $B$ sequences.

    \item \textbf{Evaluate and update.}
    Evaluate all objectives for the sampled sequences
    and update
    $\mathcal{D}_t = \mathcal{D}_{t-1}
    \cup 
    \{(x, \mathbf{y}_x) : x \in \mathcal{B}_t\}$,
    where $\mathbf{y}_x$ denotes the possibly noisy
    objective observation.
\end{enumerate}
After $T$ rounds, the campaign has used $n_0+TB$ evaluations.
We return the evaluated sequences that are nondominated
according to their observed objective vectors.
\section{Experiments}
\label{sec:experiments}

We evaluate MODS on six protein optimization tasks.
We compare our framework with three iterative optimization baselines and study how performance depends on the diffusion model, steering procedure, and steering strength.

\subsection{Experimental setup}
\label{sec:experimental_setup}

\paragraph{Tasks and evaluation oracles.}
The six tasks span different proteins and fitness objectives: five have two objectives, and one has three, as summarized in Table~\ref{tab:datasets}.
To evaluate sequences without expensive laboratory measurements, we train an oracle for each objective using existing experimental assay data.
Each fixed oracle averages the predictions of a CNN ensemble and applies a task-scaled penalty to sequences beyond the labeled mutational neighborhood.
Appendix~\ref{sec:data_collection} describes the datasets and their preprocessing.
Appendix~\ref{sec:fitness_details} describes model architecture, reports held-out predictive accuracy and defines the distance penalty.

\paragraph{MODS configurations.}
We test MODS with two diffusion models, MDLM~\citep{sahoo2024simple} and D3PM~\citep{Austin2021D3PM}, trained on a dataset of unlabeled sequences.
Each diffusion model is paired with classifier guidance (CG) and decoupled annealing posterior sampling (DAPS), yielding four MODS configurations.
In each round, each configuration trains an ensemble of multilayer perceptron surrogates using only observations acquired during the campaign, without access to the oracle training data.
Appendices~\ref{sec:prior_training} and~\ref{sec:surrogate_details} provide details on the diffusion models and surrogates, respectively.

\paragraph{Baselines.}
We compare MODS against three alternative approaches for iterative protein optimization.
ARLM + DPO fine-tunes an autoregressive model based on ProGen2-small~\citep{nijkamp2023progen2} using direct preference optimization~\citep{Rafailov2023DPO}.
Its preference pairs are constructed from observed fitness values, and a KL penalty regularizes the model toward a fixed reference.
LaMBO-2~\citep{Gruver2023Protein} couples a diffusion model and fitness
predictors through a jointly trained, shared encoder.
It uses diffusion optimized sampling (NOS) to guide generation with expected hypervolume improvement~\citep{Daulton2020Differentiable} and saliency maps to select positions for mutation.
SAGE-Prot~\citep{Lim2025SAGEProt} generates candidate pools with pretrained LSTM and transformer-decoder models, diversifies them through mutation, crossover, and homologous search, and fine-tunes the generator on high-fitness sequences.
Unlike these baselines, MODS keeps the generative model fixed throughout optimization.
Only its fitness surrogates are updated from new observations, and unlike LaMBO-2, these surrogates are decoupled from the generative model.

\paragraph{Optimization protocol and metrics.}
Each campaign begins with $n_0 = 100$ sequences sampled from the generative model, followed by eight rounds of $100$ additional evaluations. All methods therefore receive the same budget of $900$ sequences, with every objective evaluated for every sequence. We repeat each configuration five times and report means and standard errors.
We measure performance by the hypervolume of all sequences evaluated up to each round.
For each task, the reference point is the component-wise minimum across all sequences evaluated by all methods.

\paragraph{Hyperparameter selection.}
The steering parameters for CG and DAPS are chosen based on their performance on a separate set of random seeds.
For each diffusion model, the best DAPS setting achieves a higher final mean hypervolume than the best CG setting for each task.
We therefore show only the two DAPS variants in the main comparisons.
For MODS (MDLM + DAPS), the main results report the results for $\alpha=2^{12}$ and for MODS (D3PM + DAPS) for $\alpha=2^{14}$, both of which performed well across tasks.
For ARLM + DPO, we report the results for the adaptation coefficient that achieved the highest hypervolume on each task.
Appendix~\ref{sec:guidance_details} reports hyperparameter selection results and all results for both steering procedures.

\subsection{Results}
\label{sec:results}

\begin{figure*}[t]
    \centering

    \begin{subfigure}[t]{0.48\linewidth}
        \centering
        \includegraphics[width=\linewidth]{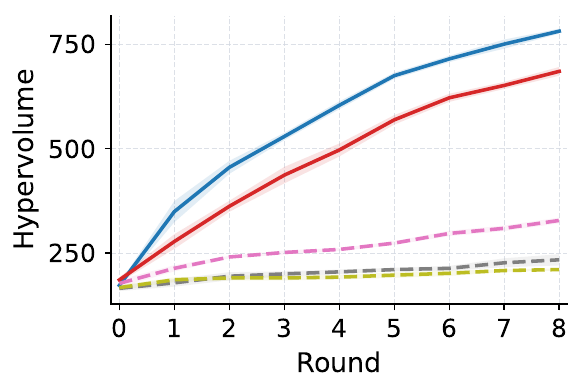}
        \caption{GB1}
    \end{subfigure}
    \hfill
    \begin{subfigure}[t]{0.48\linewidth}
        \centering
        \includegraphics[width=\linewidth]{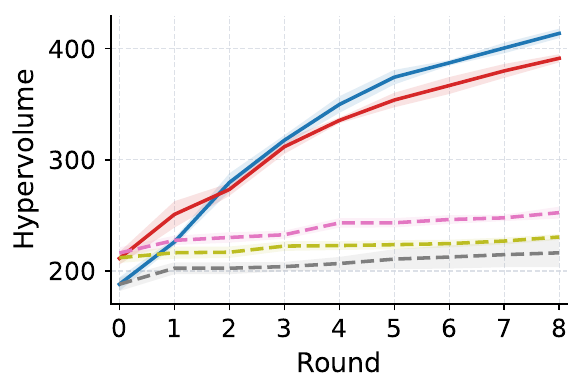}
        \caption{ERBIN-PDZ1}
    \end{subfigure}
    \\[1ex]

    \begin{subfigure}[t]{0.48\linewidth}
        \centering
        \includegraphics[width=\linewidth]{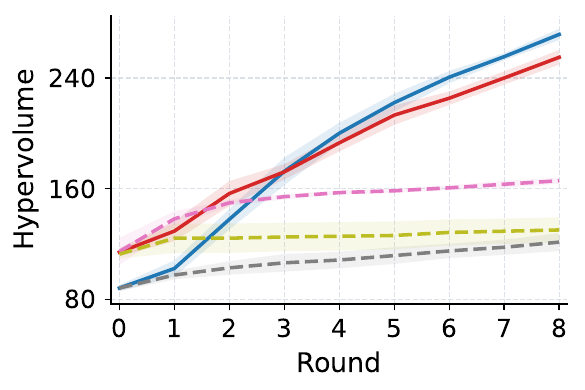}
        \caption{NHERF3-PDZ2}
    \end{subfigure}
    \hfill
    \begin{subfigure}[t]{0.48\linewidth}
        \centering
        \includegraphics[width=\linewidth]{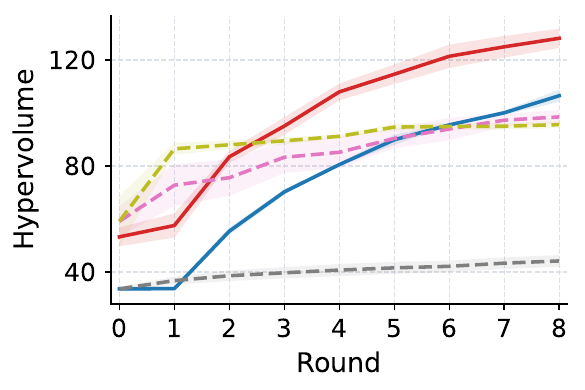}
        \caption{KRAS}
    \end{subfigure}
    \\[1ex]

    \begin{subfigure}[t]{0.48\linewidth}
        \centering
        \includegraphics[width=\linewidth]{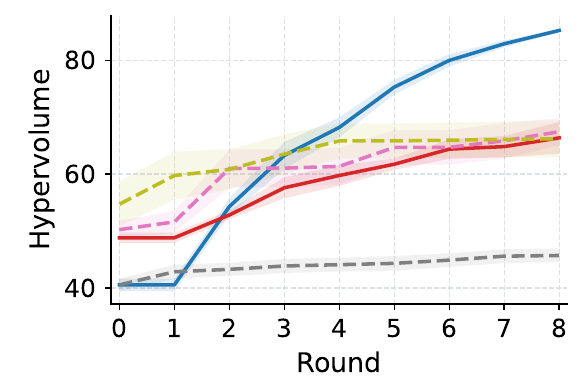}
        \caption{Src}
    \end{subfigure}
    \hfill
    \begin{subfigure}[t]{0.48\linewidth}
        \centering
        \includegraphics[width=\linewidth]{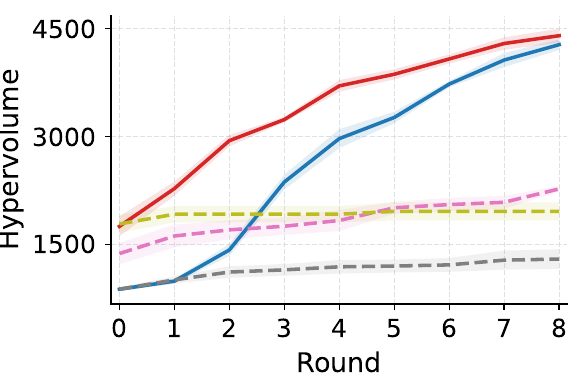}
        \caption{NHERF3-PDZ1}
    \end{subfigure}

    \vspace{0.6em}
    \includegraphics[width=0.85\linewidth]{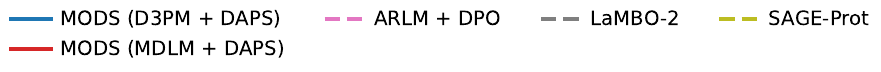}

    \caption{
        \textbf{Hypervolume over optimization rounds.}
        Hypervolume over all evaluated sequences up to each round for each task.
        Lines and shaded regions show the mean $\pm$ one standard error of the mean over five repetitions.
    }
    \label{fig:main_results}
\end{figure*}

\paragraph{Overall optimization performance.}
Figure~\ref{fig:main_results} shows the hypervolume of evaluated sequences over the optimization rounds.
We evaluate MODS with both diffusion models, D3PM and MDLM, to assess its flexibility over different generative models.
On GB1, ERBIN-PDZ1, NHERF3-PDZ2, KRAS, and NHERF3-PDZ1, both DAPS variants achieve higher final mean hypervolume than ARLM~+~DPO, LaMBO-2, and SAGE-Prot.
On Src, MODS with D3PM + DAPS achieves the highest final mean hypervolume, while MDLM + DAPS achieves $66.4 \pm 2.8$, which is close to $67.5 \pm 2.4$ for ARLM + DPO, and $66.2 \pm 3.2$ for SAGE-Prot, and clearly better than $45.7 \pm 1.2$ for LaMBO-2.
Table~\ref{tab:main_hypervolume_pvalues} reports $p$-values that compare the MODS configurations against each baseline.
Table~\ref{tab:results_hv} also shows that each DAPS variant improves over unsteered sampling from the same diffusion model under the common evaluation budget.
Together, these results demonstrate that the benefits of MODS are not limited to a particular diffusion model, while generative model's influence varies by task. Further details on the sampled sequences and the run time of MODS' components are given in Appendix~\ref{sec:additional_results}. 

\begin{table}[t]
    \centering
    \caption{
    \textbf{Hypervolume across six protein tasks.} Entries report mean $\pm$ one standard error of the mean over five repetitions. Bold identifies the highest mean among all the shown methods for each problem.
    }
    \label{tab:results_hv}
    \begingroup
        \setlength{\tabcolsep}{3pt}
        \renewcommand{\arraystretch}{0.8}
        \resizebox{\linewidth}{!}{%
        \begin{tabular}{@{}l *2{M{3.1}{2.1}} *2{M{3.1}{1.1}} M{2.1}{1.1} M{4.1}{3.1} @{}}
            \toprule
            Method & \multicolumn{1}{c}{GB1} & \multicolumn{1}{c}{\shortstack{ERBIN\\PDZ1}} & \multicolumn{1}{c}{\shortstack{NHERF3\\PDZ2}} & \multicolumn{1}{c}{KRAS} & \multicolumn{1}{c}{Src} & \multicolumn{1}{c}{\shortstack{NHERF3\\PDZ1}} \\
            \midrule
            MDLM (unsteered)
            & 231.6 \pm 11.4
            & 231.4 \pm 2.0
            & 145.4 \pm 5.0
            & 70.0 \pm 6.1
            & 64.0 \pm 2.4
            & 2244.8 \pm 78.8 \\
            \addlinespace
            
            D3PM (unsteered)
            & 225.0 \pm 4.8
            & 221.0 \pm 5.5
            & 95.7 \pm 3.4
            & 40.7 \pm 2.1
            & 44.7 \pm 1.0
            & 1004.0 \pm 28.4 \\
            \midrule
            
            MODS (MDLM + DAPS)
            & 685.3 \pm 10.4
            & 391.4 \pm 3.7
            & 254.9 \pm 5.3
            & \bfseries 128.2 \pm 3.6 
            & 66.4 \pm 2.8 
            & \bfseries 4403.5 \pm 97.0\\
            \addlinespace
            
            MODS (D3PM + DAPS)
            & \bfseries 781.4 \pm 5.9 
            & \bfseries 413.8 \pm 4.2 
            & \bfseries 271.5 \pm 4.1 
            & 106.5 \pm  2.3 
            & \bfseries 85.2 \pm 0.3 
            & 4278.9 \pm 72.5 \\
            \midrule
            
            ARLM + DPO
            & 328.4 \pm 6.8
            & 252.4 \pm 5.3
            & 165.7 \pm 2.1
            & 98.5 \pm 2.5
            & 67.5 \pm 2.4
            & 2272.7 \pm 39.7 \\
            \addlinespace
            
            SAGE-Prot
            & 210.8 \pm 2.5
            & 230.5 \pm 2.6
            & 130.2 \pm 9.1
            & 95.6 \pm 1.3
            & 66.2 \pm 3.2
            & 1960.6 \pm 126.3 \\
            \addlinespace
            
            LaMBO-2
            & 234.2 \pm 13.5
            & 216.2 \pm 12.8
            & 121.4 \pm 6.4
            & 44.1 \pm 2.2
            & 45.7 \pm 1.2
            & 1296.8 \pm 136.3 \\
            \bottomrule
        \end{tabular}%
        }
    \endgroup
\end{table}

\begin{figure*}[t]
    \centering

    \begin{subfigure}[t]{0.32\linewidth}
        \centering
        \includegraphics[width=\linewidth]{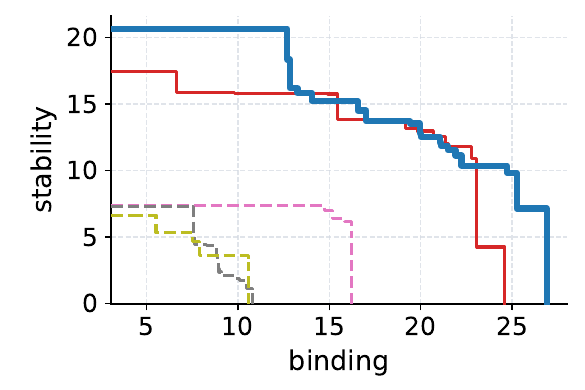}
        \caption{GB1}
    \end{subfigure}
    \hfill
    \begin{subfigure}[t]{0.32\linewidth}
        \centering
        \includegraphics[width=\linewidth]{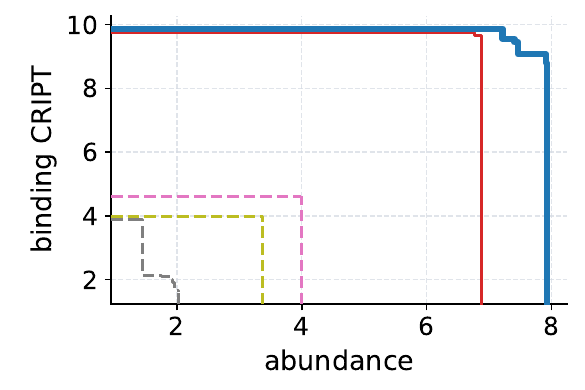}
        \caption{ERBIN-PDZ1}
    \end{subfigure}
    \hfill
    \begin{subfigure}[t]{0.32\linewidth}
        \centering
        \includegraphics[width=\linewidth]{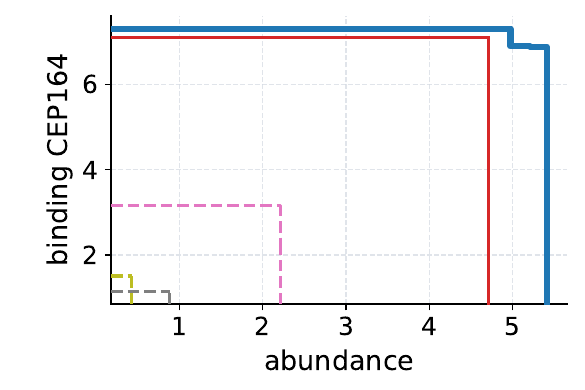}
        \caption{NHERF3-PDZ2}
    \end{subfigure}
    \\[1ex]

    \hspace*{\fill}
    \begin{subfigure}[t]{0.32\linewidth}
        \centering
        \includegraphics[width=\linewidth]{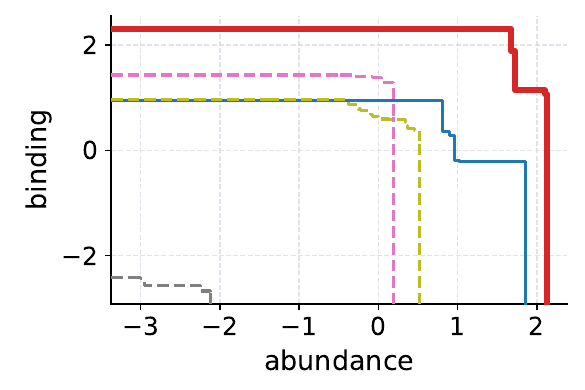}
        \caption{KRAS}
    \end{subfigure}
    \hfill
    \begin{subfigure}[t]{0.32\linewidth}
        \centering
        \includegraphics[width=\linewidth]{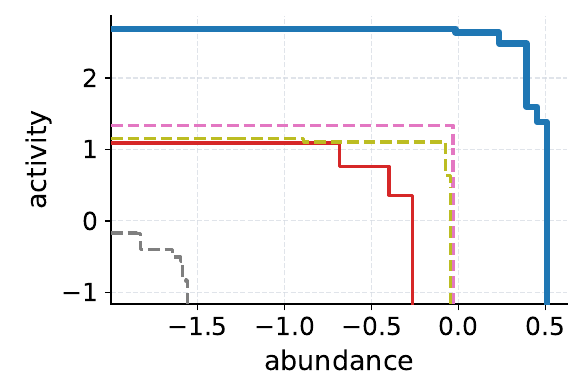}
        \caption{Src}
    \end{subfigure}
    \hspace*{\fill}

    \vspace{0.6em}
    \includegraphics[width=0.9\linewidth]{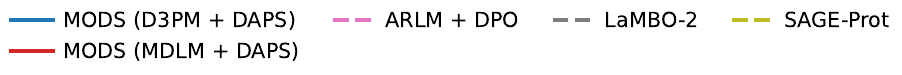}

    \caption{
        \textbf{Final Pareto fronts.}
        Each panel shows the nondominated objective vectors after the final evaluation round.
        For each method, we plot the repetition with median final hypervolume; the front with the largest hypervolume is highlighted in bold.
    }
    \label{fig:pareto-comparison}
\end{figure*}

\paragraph{Coverage of objective tradeoffs.}
For the five two-objective tasks, Figure~\ref{fig:pareto-comparison} provides a qualitative view of the final tradeoffs recovered by each method.
The three-objective NHERF3-PDZ1 task is omitted because its Pareto front cannot be represented by a single two-dimensional plot.
The differences are most visible on GB1, ERBIN-PDZ1, KRAS, and Src, where the highest-hypervolume fronts extend farther along complementary objective directions.

\paragraph{Effect of steering strength.}
Figures~\ref{fig:ablation-daps} and~\ref{fig:ablation-cg} show how the KL-regularization coefficient $\alpha^{-1}$ in Equation~\ref{eq:mods_regularized_objective} affects optimization performance.
Reducing $\alpha^{-1}$ strengthens steering by allowing the sampling distributions $q_t$ to deviate more from the pretrained diffusion model.
With weak regularization, sampling relies more heavily on surrogate predictions far from the observed data, where those predictions may be unreliable.
The resulting performance decline in some settings is consistent with oversteering, although we do not directly test this mechanism.
For the MDLM model, performance peaks at an intermediate steering strength on KRAS and Src with DAPS and on NHERF3-PDZ2 and KRAS with CG.
Further strengthening steering reduces final hypervolume in these settings.
In most other diffusion model--task combinations, stronger steering improves performance or eventually reaches a plateau.
These results show that the appropriate steering strength depends jointly on the prior, steering procedure, and task.


\section{Conclusion}
\label{sec:conclusion}

We introduced MODS, a modular framework for multi-objective protein optimization with limited feedback. MODS samples augmented Chebyshev scalarization weights and surrogate ensemble members to construct rewards targeting different objective tradeoffs. These rewards provide a modular interface between multi-objective acquisition and generative steering: they can be used with compatible inference-time steering procedures without modifying or retraining the generative model. As new observations are collected, MODS updates the fitness surrogates and redirects generation while keeping the generative model fixed.

We evaluated MODS with two diffusion models and two steering procedures across six protein optimization tasks. MODS performed strongly with both diffusion models, although performance depended on the choice of generative model and steering procedure. 
For each diffusion model, MODS either achieved the highest mean final hypervolume or performed similar to the evaluated baselines. 
These results show that MODS can discover strong and diverse objective tradeoffs within a few rounds of batched feedback while supporting different model--steering combinations.

\paragraph{Scope and limitations.}
Our evaluation uses predictive oracles trained and validated on experimental assay data, enabling controlled comparisons across tasks and methods.
The oracles use a distance-based penalty to discourage exploration beyond labeled mutational neighborhoods, where prediction accuracy is uncertain.
Laboratory studies remain necessary to determine whether observed gains translate to experimentally measured fitness.
We evaluated MODS on a limited set of protein tasks, diffusion priors, and steering procedures. Experiments across more protein families, generative models, and steering procedures would help establish the framework's generality.

\paragraph{Future work.}
Beyond laboratory validation, a natural next step is to combine feedback at different fidelities, such as inexpensive computational predictions, screening assays, and more costly measurements. Doing so would require deciding both which sequences to evaluate and at what fidelity. Another promising direction is to transfer fitness information across related proteins, targets, or assay conditions. Extending reuse from generative sequence models to task-specific fitness models could further reduce the number of measurements required for new optimization campaigns.

\subsection*{AI use statement}
In this work, we used generative AI tools for 
providing feedback on research methodology or experiments, 
and 
assisting with translation.
We have not used generative AI tools to 
generate synthetic data sets,
help develop theoretical models or conceptual frameworks,
propose or refine hypotheses,
implement methods,
clean and reformat dataset,
support qualitative and thematic data analysis,
interpret results,
and
formulate mathematical claims,
provide critical ingredients for proving mathematical claims,
assist in the writing of proofs
are not applicable to this work.
Additionally, we used generative AI tools for 
editing non-methodological parts of software code,
sourcing/searching for information,
and
editing the research paper to improve readability. 
We have reviewed all AI-assisted work.
We checked, verified and tested all LLM-generated outputs including text, code, summaries, analysis and edits before incorporating them into work.
We take responsibility for the final content of this work,
including text, claims or artifacts produced with the aid of generative AI.

\subsection*{Ethics statement}

Our study does not involve human subjects, personally identifiable information, or sensitive personal data. 
All datasets used are publicly available, and we comply with their original licenses and terms of use.
The pretrained generative-prior and fitness-oracle checkpoints are trained solely on these datasets, no private or proprietary user data were used.
Like other generative-modeling research, it may inherit general risks associated with the generation of misleading, biased, or otherwise harmful content.
Because our work involves protein fitness prediction and generative design, we considered biosafety and dual-use concerns.
We do not intentionally generate or release protein sequences with known harmful function.
This work is intended for methodological research on optimization with learned generative models and oracles.

\subsection*{Reproducibility statement}
Dataset sources and processing, model architectures, optimization settings, evaluation protocols, checkpoint selection, and additional details are documented in Section~\ref{sec:experimental_setup} and Appendices ~\ref{sec:data_collection}, ~\ref{sec:prior_training}, ~\ref{sec:fitness_details}, and ~\ref{sec:surrogate_details}.
We intend to release the source code, processed datasets, data splits, pretrained generative-prior and fitness-oracle checkpoints, experiment configurations, random seeds, environment setup instructions, saved experimental outputs, and analysis scripts that support regeneration of the reported tables and figures.
At the time of submission, these are not yet publicly available. 
An anonymized reproducibility package will be made available to reviewers during the rebuttal.

\bibliography{iclr2027_conference}
\bibliographystyle{iclr2027_conference}

\clearpage

\appendix
\counterwithin{figure}{section}
\counterwithin{table}{section}

\section{Datasets and preprocessing}
\label{sec:data_collection}
\begin{table}[t]
    \caption{\textbf{Summary of used datasets.} Length $L$ of the sequences, multiple sequence alignment (MSA) counts in the training datasets for the generative model, number of sequences in the train and test datasets per objective for the oracle models.}
    \label{tab:datasets}
    \centering
    \vspace{6pt}
    \begingroup
        \setlength{\tabcolsep}{2.2pt}
        \renewcommand{\arraystretch}{1.08}
        \renewcommand{\tabularxcolumn}[1]{m{#1}}
            \begin{tabularx}{\linewidth}{@{} l r >{\raggedright\arraybackslash}m{0.22\linewidth} rrr >{\raggedright\arraybackslash}X @{}}
            \toprule
            Dataset 
            & Length 
            & Objectives
            & \shortstack{MSA\\size} 
            & \shortstack{Train\\fitness} 
            & \shortstack{Test\\fitness} 
            & Reference \\
            \midrule
            GB1 
            & 56 
            & binding \newline stability$^{\dagger}$
            & 128,881 
            & \begin{tabular}{@{}r@{}}385,860\\382,946\end{tabular} 
            & \begin{tabular}{@{}r@{}}96,465\\95,693\end{tabular} 
            & \citet{olson2014gb1}, \newline \citet{wu2016stability} \\
            \midrule
            KRAS 
            & 188 
            & abundance \newline DARPin K55 binding 
            & 105,807 
            & 17,035 
            & 4,025 
            & \citet{weng2024kras} \\
            \midrule
            ERBIN-PDZ1 
            & 90 
            & abundance \newline CRIPT binding 
            & 71,641 
            & 26,996 
            & 3,375 
            & \citet{martiaranda2026pdz} \\
            \midrule
            NHERF3-PDZ2 
            & 111 
            & abundance \newline CEP164 binding 
            & 39,809 
            & 30,761 
            & 3,847 
            & \citet{martiaranda2026pdz} \\
            \midrule
            Src$^{\ddagger}$ 
            & 269 
            & abundance \newline kinase activity 
            & 753,770 
            & 41,599 
            & 5,201 
            & \citet{beltran2026src} \\
            \midrule
            NHERF3-PDZ1
            & 106
            & abundance \newline DAP1 binding \newline UHMK binding
            & 63,450
            & 24,971
            & 3,121
            & \citet{martiaranda2026pdz} \\
            \bottomrule
        \end{tabularx}
        \par\medskip
        \begin{minipage}{\linewidth}
            $^{\dagger}$ GB1 stability is inferred from single-mutant estimates, summed for double mutants.\\
            $^{\ddagger}$ Src models residues 268--536; assays used the full-length protein.
        \end{minipage}
    \endgroup
\end{table}

For each protein task, we use two complementary datasets that differ in purpose.
The first is an unlabeled homolog dataset gathered from UniRef90~\citep{suzek2007uniref}, providing a natural-sequence distribution to train discrete diffusion models.
The second is a labeled sequence--fitness dataset, obtained from mutational studies, which is used to train the prediction oracles.
These two data sources are kept separate throughout: no fitness labels are used during diffusion training, and no homolog data are treated as fitness labels.
An overview of the used data is represented in Table~\ref{tab:datasets}.

For each task, the datasets for the diffusion models are obtained by searching UniRef90~\citep{suzek2007uniref} for homologs of a parent sequence using two iterations of JackHMMER~\citep{johnson2010jackhmmer}.
The resulting multiple sequence alignments (MSAs) are processed by removing insertions relative to the parent and replacing alignment gaps with the corresponding parent residues. A homolog is retained only if it covers at least 75\% of the parent sequence without gaps prior to gap filling. The processed sequences form fixed-length pseudo-natural sequences that are used to train the generative model.
To reduce redundancy among homologous sequences, we cluster them using MMseqs2~\citep{steinegger2017mmseqs2} at 80\% sequence identity, i.e., sequences sharing at least 80\% identical residues are grouped into the same cluster. We hold out 5\% of clusters and their associated sequences as a validation set. The remaining sequences are resampled with replacement using weights proportional to $(1 + \log n_c)^{-1}$, where $n_c$ is the cluster size.
This reduces the contribution of large, closely related clusters. Because the MSA for GB1 contains relatively few homologs, we augment each homolog with all single amino acid substitutions.

The fitness datasets are obtained from mutational studies listed in Table~\ref{tab:datasets}.
They contain single and double amino acid mutations with associated fitness values. For GB1, the stability objective is constructed from published single-mutant stability estimates~\citep{wu2016stability}, by augmenting the training data with double-mutant stability values obtained by summing the corresponding single-mutant effects. 
For other proteins, the fitness values are taken directly from the corresponding mutational studies. These sequence-fitness pairs are used as labeled data to train the oracle models to serve as a proxy for wet-lab assay measurements, allowing generated variants to be evaluated in silico.

\section{Diffusion models and sample quality}
\label{sec:prior_training}

\begin{table}[t]
    \centering
    \caption{\textbf{Generative models.}}
    \label{tab:generative_priors}
    \setlength{\tabcolsep}{2.5pt}
    \renewcommand{\arraystretch}{1.15}
    \begin{tabularx}{\linewidth}{@{}lll r X@{}}
        \toprule
        Model
        & Family
        & Corruption
        & \shortstack[l]{Trainable\\parameters}
        & Architecture and initialization \\
        \midrule
        D3PM 
        & Discrete diffusion 
        & Uniform 
        & 37.9\,M
        & ByteNet denoiser finetuned from \mbox{EvoDiff} 38M-Uniform \citep{alamdari2023evodiff}. \\
        MDLM 
        & Discrete diffusion 
        & Absorbing state 
        & 28.6\,M
        & Transformer denoiser \citep{sahoo2024simple}; trained from scratch. \\
        \bottomrule
    \end{tabularx}
\end{table}

Table~\ref{tab:generative_priors} lists the model sizes, architectures and initialization for the generative models, used in their training.

\begin{figure*}[t]
    \centering
    \begin{subfigure}[t]{0.48\linewidth}
        \centering
        \includegraphics[width=\linewidth]{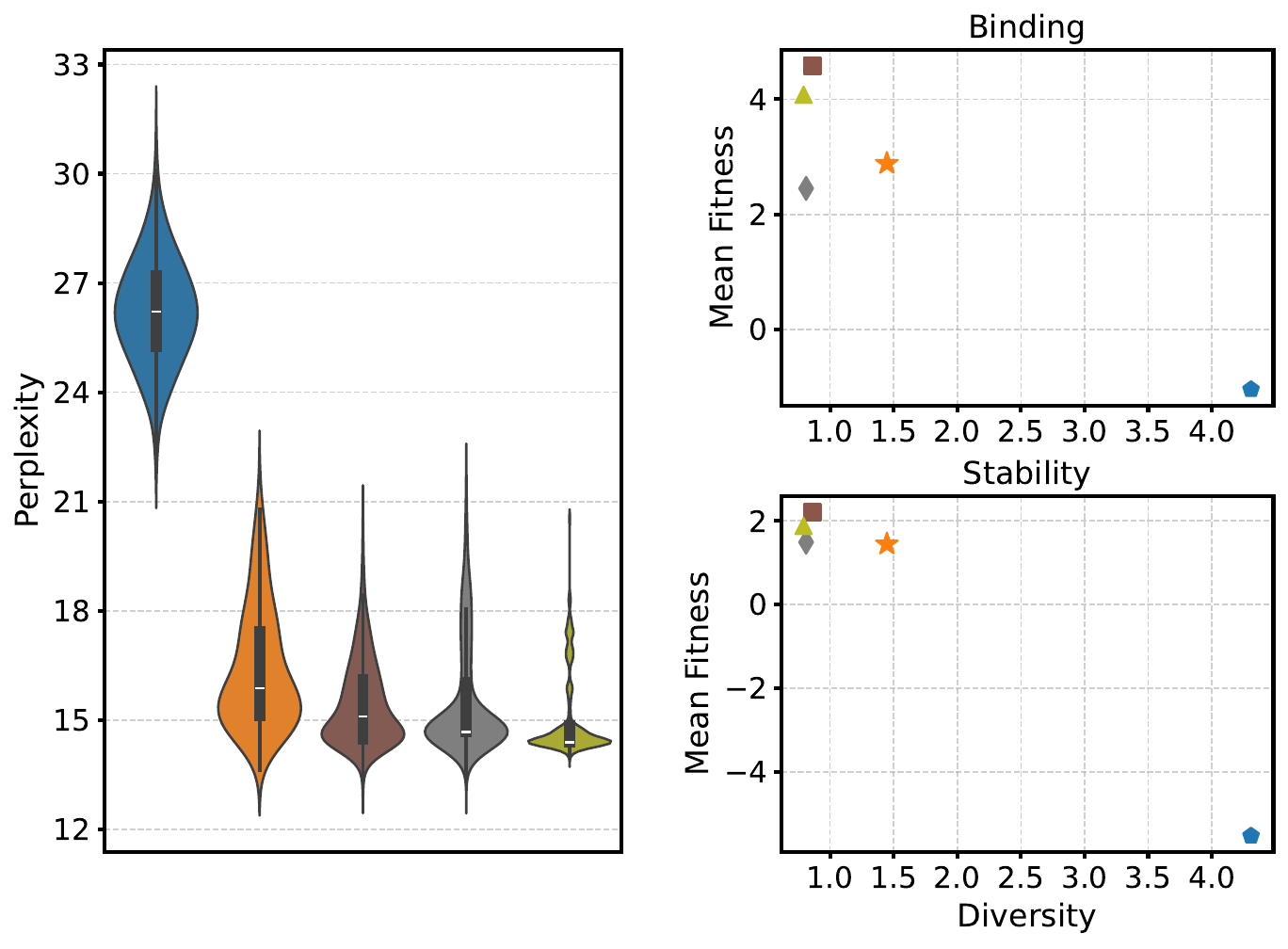}
        \caption{GB1}
    \end{subfigure}
    \hfill
    \begin{subfigure}[t]{0.48\linewidth}
        \centering
        \includegraphics[width=\linewidth]{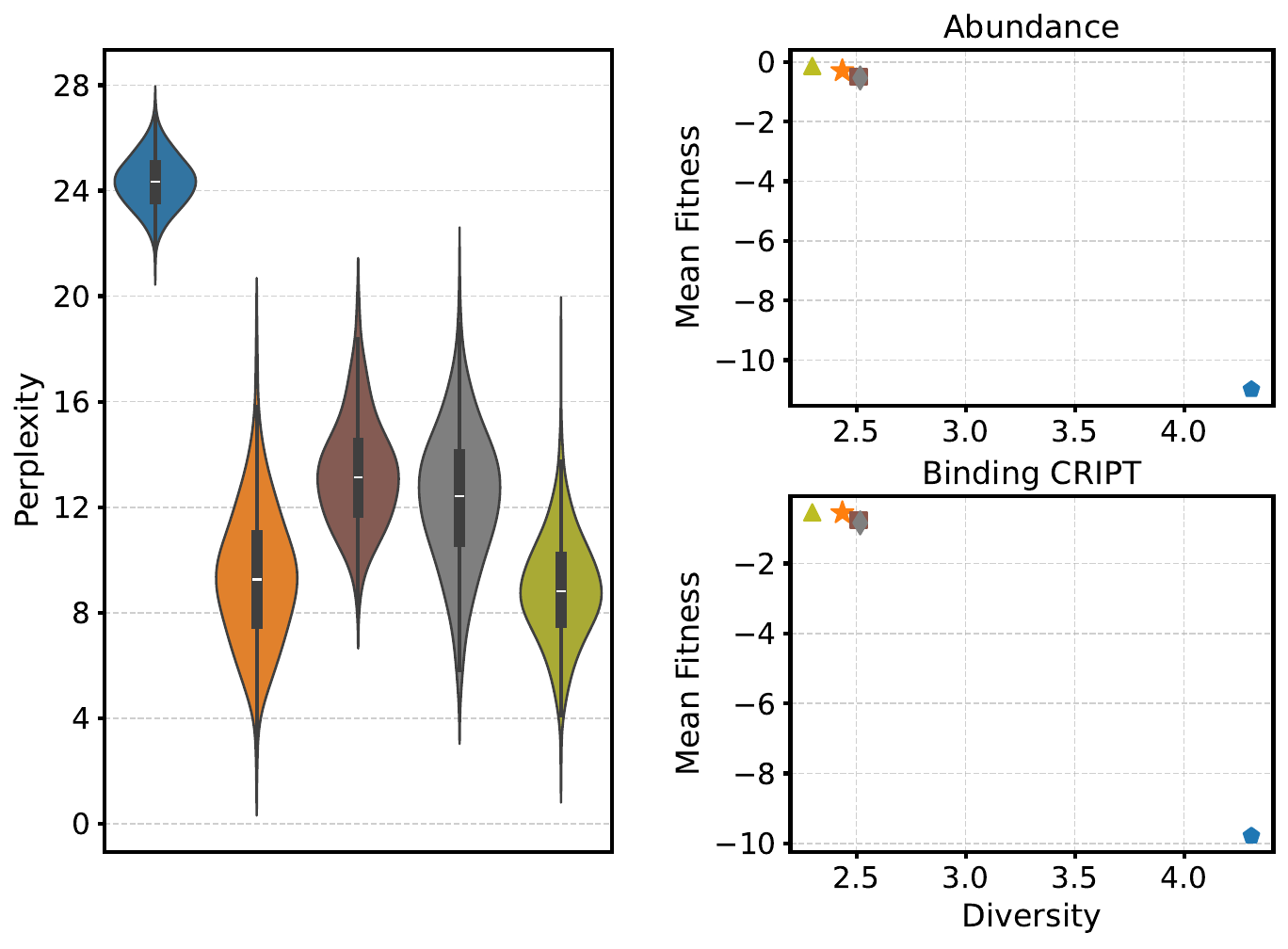}
        \caption{ERBIN-PDZ1}
    \end{subfigure}
    \vspace{0.6em}
    \begin{subfigure}[t]{0.48\linewidth}
        \centering
        \includegraphics[width=\linewidth]{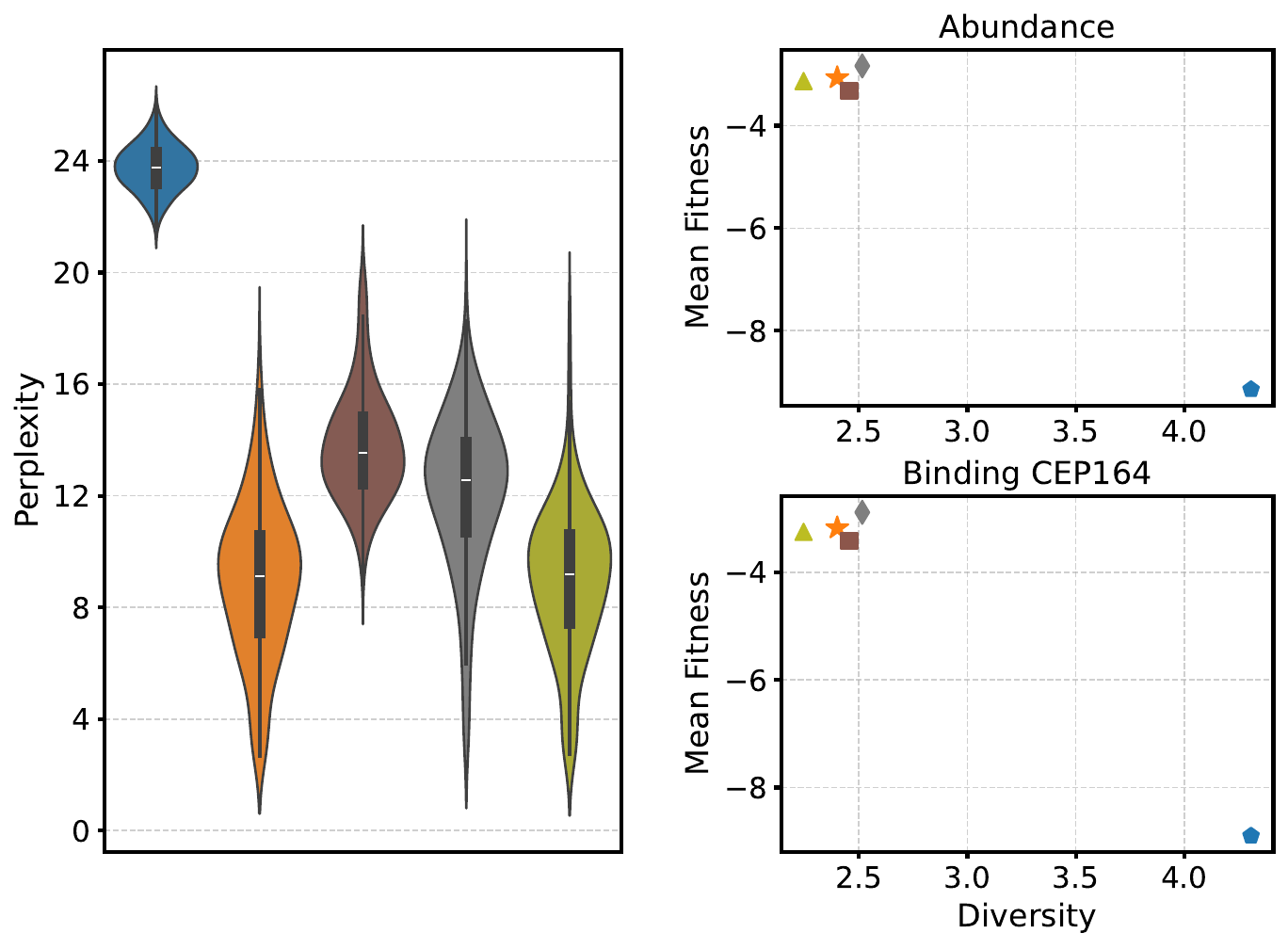}
        \caption{NHERF3-PDZ2}
    \end{subfigure}
    \hfill
    \begin{subfigure}[t]{0.48\linewidth}
        \centering
        \includegraphics[width=\linewidth]{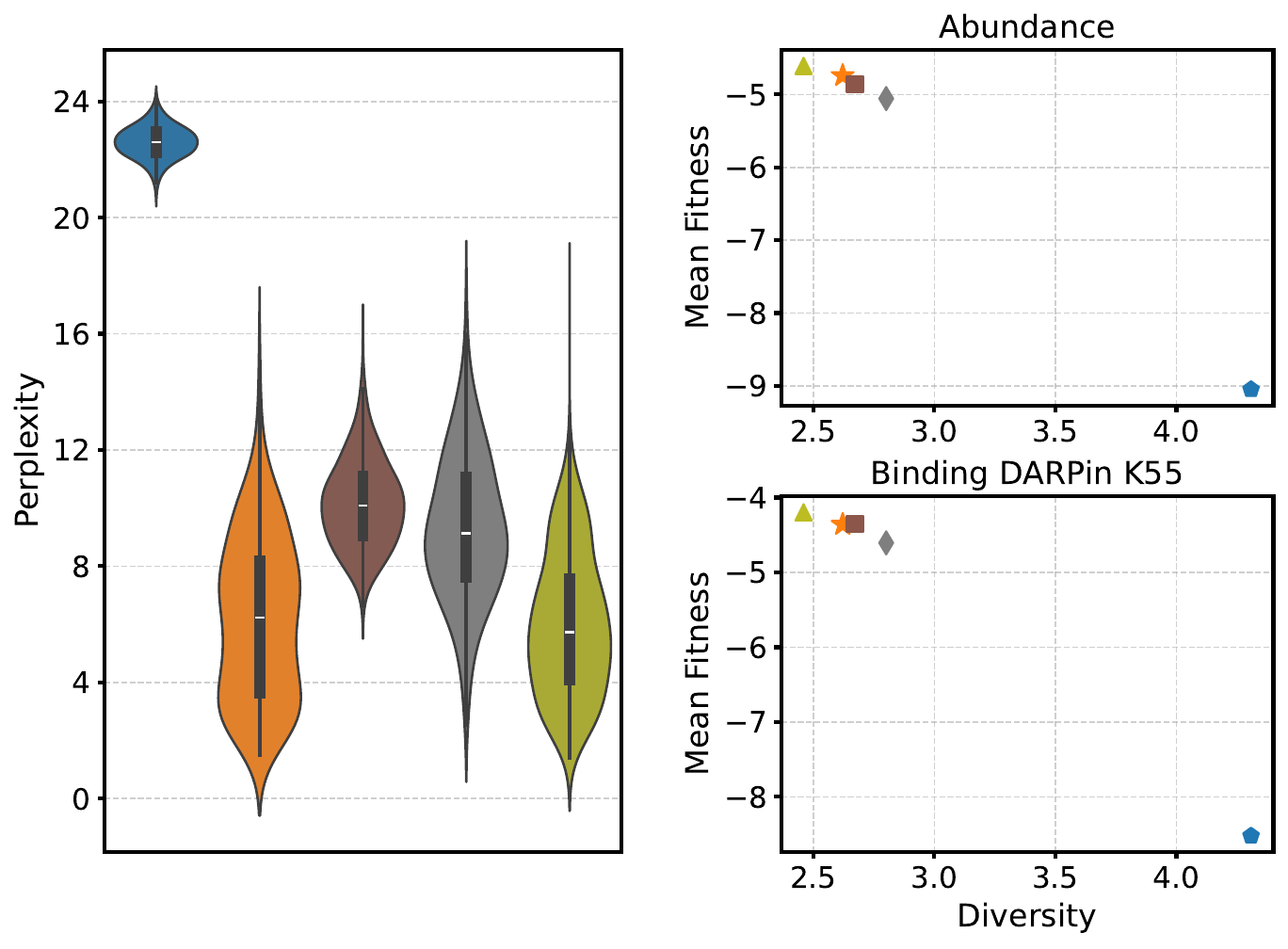}
        \caption{KRAS}
    \end{subfigure}
    \vspace{0.6em}
    \begin{subfigure}[t]{0.48\linewidth}
        \centering
        \includegraphics[width=\linewidth]{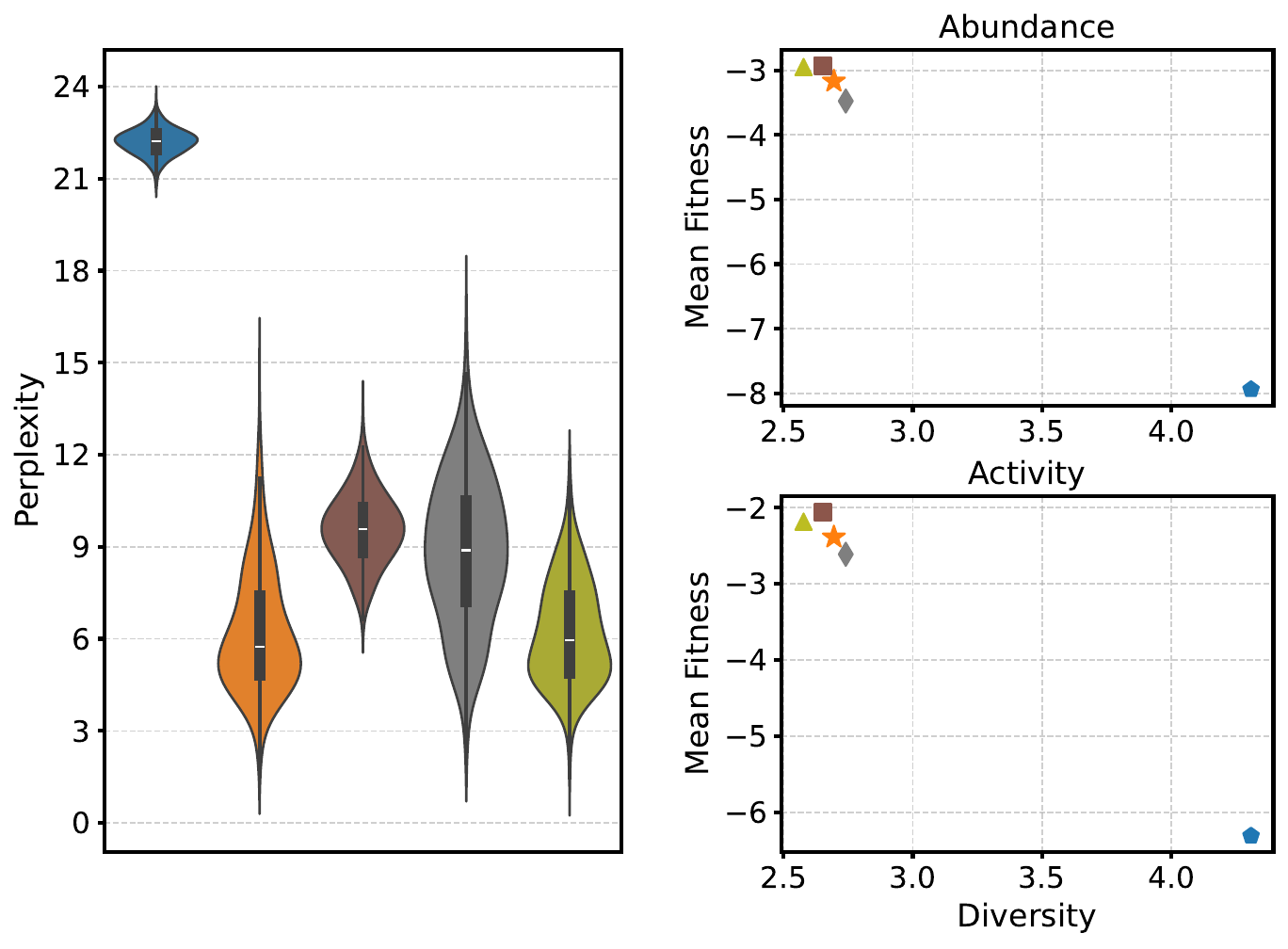}
        \caption{Src}
    \end{subfigure}
    \hfill
    \begin{subfigure}[t]{0.48\linewidth}
        \centering
        \includegraphics[width=\linewidth]{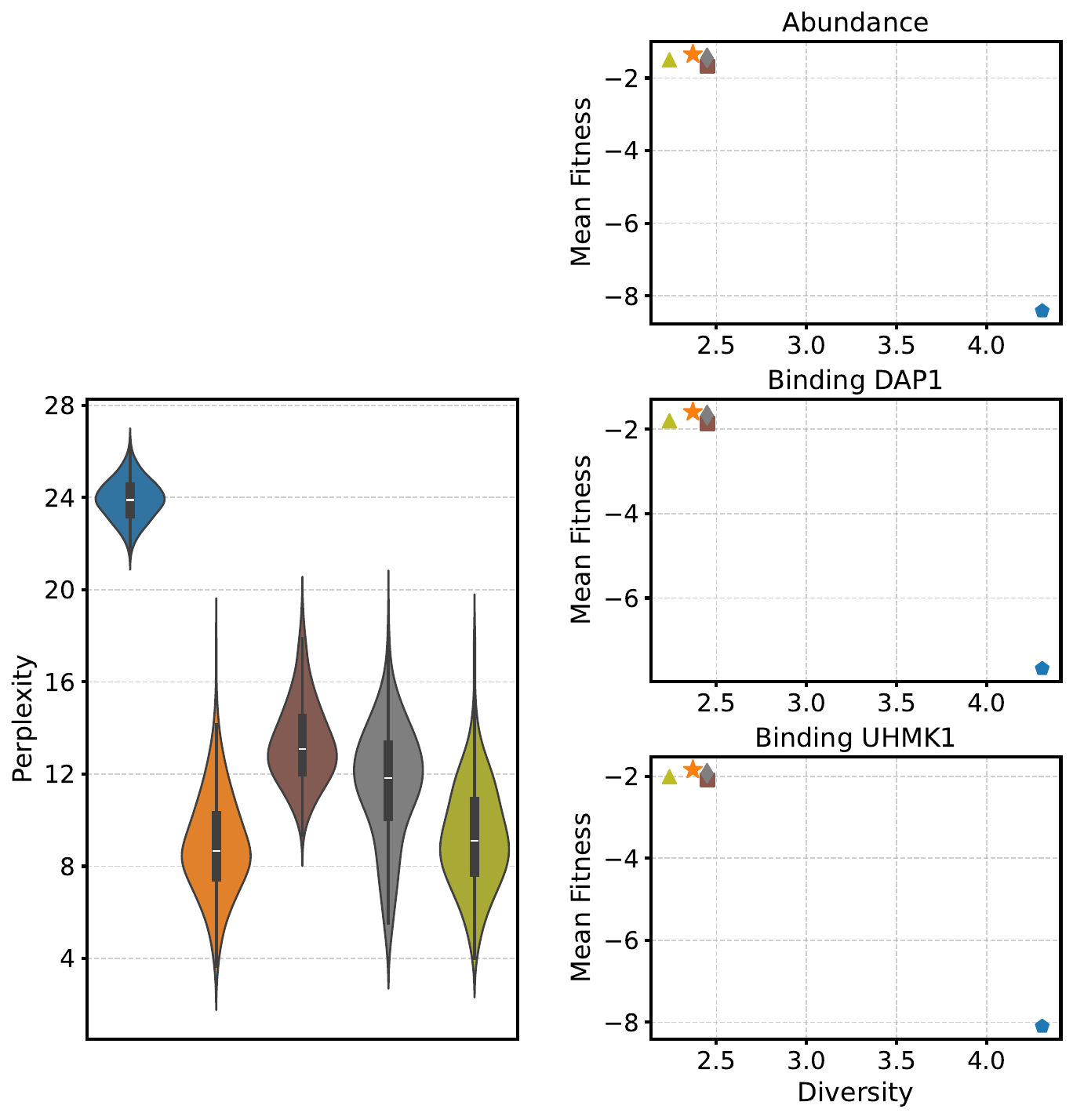}
        \caption{NHERF3-PDZ1}
    \end{subfigure}
    \vspace{0.4em}
    \includegraphics[width=0.55\linewidth]{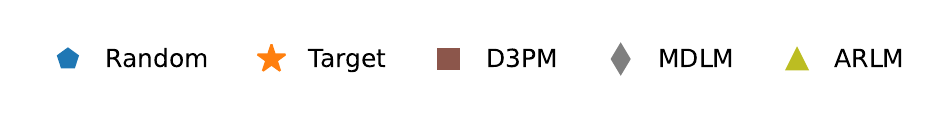}    
    \caption{\textbf{Generative model evaluation.} Perplexity distributions are shown on the left. The right panels report the mean oracle fitness versus sequence diversity for all objectives.}
    \label{fig:all_priors}
\end{figure*}

Figure~\ref{fig:all_priors} summarizes properties of sequences drawn from each diffusion model for each protein task.
The violin plots compare the perplexity of the training data with the sequences generated by the models, uniformly random sequences, and the training sequences.
Perplexities are calculated with respect to the \qty{764}{M}-parameter ProGen2-base model~\citep{nijkamp2023progen2}.
A sequence with low perplexity corresponds to a high likelihood in this model.
This allows comparing the distributions of the learned models with the target training data.
The scatter plots show the mean oracle fitness against the sequence diversity of the generated samples.
Sequence diversity is the average Shannon entropy of the amino acid distribution generated at each sequence position. 

Across all datasets, the random samples consistently yield the highest perplexity.
In contrast, the natural MSA sequences (``Target'') that were used to train the generative models have much lower perplexity.
The learned generative models' perplexities are clearly different from those of random sequences and better approximate the training data's distribution.

The right panels show the trade-off between the diversity of the generated sequences and their mean fitness in each objective.
Random samples achieve high diversity but have the lowest mean predicted fitness, whereas the target lies in the high-fitness, low-diversity corner. This pattern is preserved across all six protein groups.

\section{Fitness oracle evaluation and validation}
\label{sec:fitness_details}
\begin{figure}[t]
    \centering
    \includegraphics[
        width=\textwidth,
        height=0.86\textheight,
        keepaspectratio
    ]{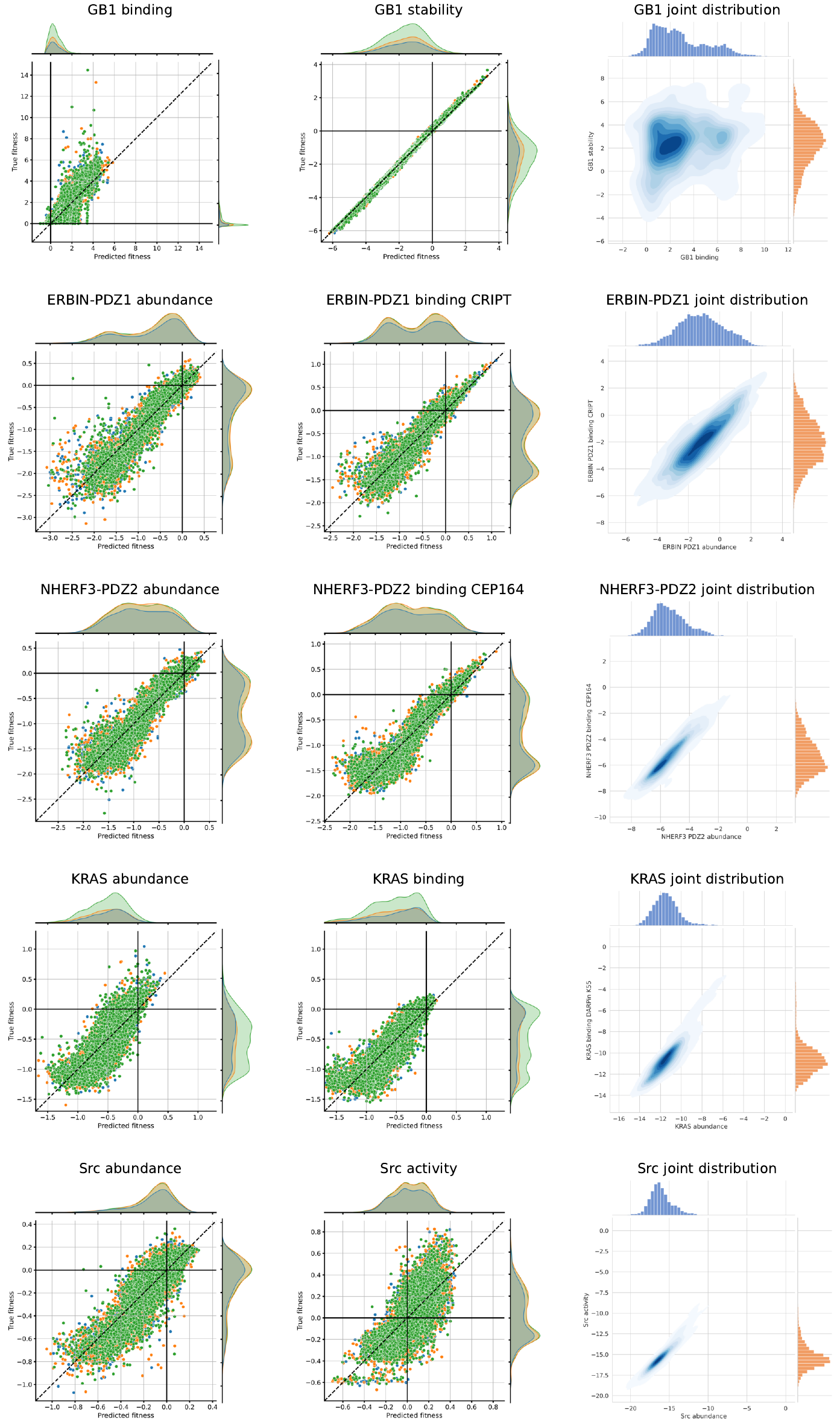}
    \includegraphics[width=0.3\linewidth]{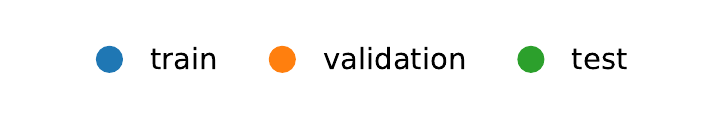}
    \vspace{-1ex}
    \caption{
        \textbf{Oracle accuracy and joint oracle distributions for two-objective tasks.} The first two columns compare predicted and true fitness for each objective. The third column shows kernel density estimates of joint oracle values for the training sequences, with marginal histograms. Darker blue indicates higher density.
    }
    \label{fig:oracle-overview}
\end{figure}

\begin{figure}[t]
    \centering
    \includegraphics[
        width=\textwidth,
        keepaspectratio
    ]{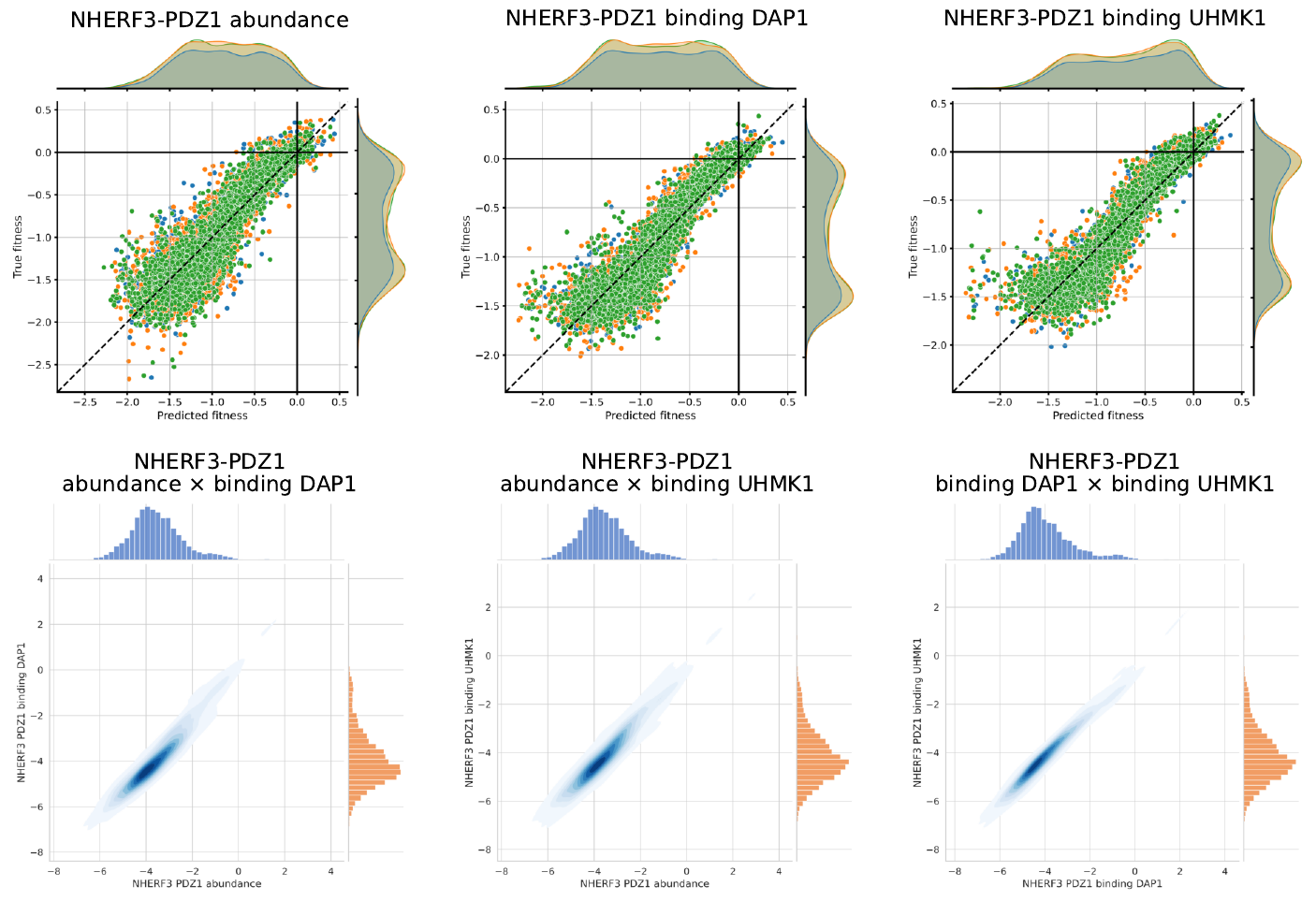}
    \includegraphics[width=0.3\linewidth]{figures/oracles/split_legend.pdf}
    \vspace{-1ex}
    \caption{
        \textbf{Oracle accuracy and joint oracle distributions for three-objective task.} The first two columns compare predicted and true fitness for each objective. The third column shows kernel density estimates of joint oracle values for the training sequences, with marginal histograms. Darker blue indicates higher density.
    }
    \label{fig:oracle-nherf3-pdz1-overview}
\end{figure}

To train the oracle models, we use an ensemble of ten one-dimensional convolutional neural networks (CNNs) to predict the objectives.
CNNs have been shown to have better extrapolation behavior than fully connected networks~\citep{Freschlin2024Neural,Sandhu2025Investigating} 
Each network takes an one-hot encoded amino acid sequence as input and comprises two convolutional layers, each followed by batch normalization and GELU activation. A final linear regression head is applied to the flattened features of the previous layers.
All oracles use $64$ and $32$ channels, respectively, with kernel size $7$.
The ensemble members are trained with different random seeds using Adam and a mean squared error loss.
The oracle's prediction $\bar f_m(x)$ is the average of all ensemble members' predictions.
Table~\ref{tab:oracle_accuracy} reports the Pearson correlations on test data splits for the resulting CNN ensembles.
Figures~\ref{fig:oracle-overview} and~\ref{fig:oracle-nherf3-pdz1-overview} show the correlation between the data and the oracle predictions on the MSA sequences that were used to train the generative models. 

\begin{table}[t]
    \centering
    \caption{\textbf{Fitness oracle accuracy.} Pearson correlations $r$ on the test set for each oracle.}
    \label{tab:oracle_accuracy}
    \begin{tabular}{@{}llclclc@{}}
        \toprule
        Protein & Objective 1 & $r$ & Objective 2 & $r$ & Objective 3 & $r$ \\
        \midrule
        GB1 & Binding & 0.89 & Inferred stability & 1.00 & -- & -- \\
        ERBIN-PDZ1 & Abundance & 0.94 & CRIPT binding & 0.95 & -- & -- \\
        NHERF3-PDZ2 & Abundance & 0.94 & CEP164 binding & 0.94 & -- & -- \\
        KRAS & Abundance & 0.83 & DARPin K55 binding & 0.89 & -- & -- \\
        Src & Abundance & 0.87 & Kinase activity & 0.76 & -- & -- \\
        NHERF3-PDZ1 & Abundance & 0.91 & DAP1 binding & 0.93 & UHMK1 binding & 0.94 \\
        \bottomrule
    \end{tabular}
\end{table}

Since the training data for the oracle only come from a limited neighborhood around a reference sequence, it has to extrapolate for more distant sequences, where predictions become increasingly unreliable.
To discourage optimizing in biologically uninformed regions, we add a penalty to the raw oracle predictions:
\begin{equation*}
    f_m(x) = \bar f_m(x) - \lambda_m \max(0, d_{\mathrm H}(x, x_{\mathrm{ref}}) - d_0).
\end{equation*}
Here, $d_{\mathrm H}$ is the Hamming distance from a reference sequence $x_{\mathrm{ref}}$, and $d_0$ defines a reference distance in which predictions remain unchanged before the penalty is applied.
The parameter $\lambda_m \geq 0$ controls the strength of the penalty for the objective $f_m$.
In our experiments, we set $d_0=4$ for GB1, ERBIN-PDZ1, NHERF3-PDZ2 and NHERF3-PDZ1, which have shorter sequences, and $d_0=8$ for KRAS and Src, which have longer sequences.
We set $\lambda_m$ to the standard deviation of the true fitness values for objective $f_m$ in the training data. This makes the penalty comparable to the spread of the training labels, so the penalization strength adjusts to objectives with different fitness scales.

\section{Surrogate architecture and training}
\label{sec:surrogate_details}
The surrogates are trained on the evaluations collected during the optimization campaign.
The labeled data used to train the oracles is not used to train surrogates.

Each surrogate is an MLP predicting each objective.
The MLP takes one-hot-encoded amino acid sequences as input, embeds them into a $ 64$-dimensional space, and flattens the features.
For CG-steering, these features are concatenated with a $64$-dimensional time embedding constructed from fixed Gaussian Fourier features, with frequency standard deviation $30$, followed by a linear layer of the same output size.
The features are then passed through three fully connected layers of width $64$, with ReLU activations between them.
A regression head then applies a further width-$64$ linear layer, a ReLU, and a final linear layer to produce the objective predictions. 

For CG, the surrogate training samples a diffusion timestep uniformly for each training data, and the diffusion model's forward corruption process is applied to the sequence. The surrogate receives the corrupted sequence and its timestep, while the regression target remains the oracle objective vector of the original clean sequence. This provides training examples at different noise levels without additional oracle queries.
For DAPS, the surrogates are trained on clean sequences.

In each optimization round, we initialize an ensemble of ten surrogates with different random seeds and train each member from scratch on all observations accumulated within the campaign. 
Training minimizes the mean squared error jointly over all objectives, using Adam with a learning rate of $10^{-3}$ and a shuffled batch size of $20$. Each member is trained for $1000$ epochs for CG and $200$ epochs for DAPS.

\section{Hyperparameter and steering sensitivity}
\label{sec:guidance_details}

All experiments use $\rho=0.05$ for the augmented Chebyshev scalarization.

\begin{table}
    \centering
    \caption{
    \textbf{Final hypervolume across all evaluated methods for hyperparameter selection.} Entries report the mean over two repetitions. Bold indicates the highest mean among different parameters of a method. Green highlights indicates the highest mean among all methods.
    }
    \label{tab:hypervolume_hpo}
    \begingroup
        \setlength{\tabcolsep}{4pt}
        \renewcommand{\arraystretch}{1.15}
        \setlastcellformat{S[table-format=4.1, mode=text, reset-text-series=false]}
        \begin{tabular}{@{}l *4{S[table-format=3.1, mode=text, reset-text-series=false]} S[table-format=2.1, mode=text, reset-text-series=false] S[table-format=4.1, mode=text, reset-text-series=false] @{}}
            \toprule
            Setting & {GB1} & \shortstack{ERBIN\\PDZ1} & \shortstack{NHERF3\\PDZ2} & {KRAS} & {Src} & \shortstack{NHERF3\\PDZ1} \\
            \midrule
            \multicolumn{7}{@{}l@{}}{\textbf{MODS (MDLM + CG)}} \\[1pt]
            $\tau=0.4^7$ & \bfseries 485.9 & \bfseries 320.9 & 216.2 & 101.5 & 74.8 & 3757.8 \\
            $\tau=0.4^6$ & 466.2 & 312.4 & \bfseries 227.8 & 107.7 & \bfseries 82.3 & \bfseries 3764.8 \\
            $\tau=0.4^5$ & 425.2 & 297.3 & 212.5 & \bfseries 108.3 & 76.9 & 3251.6 \\
            $\tau=0.4^4$ & 349.4 & 260.4 & 190.9 & 99.9 & 74.8 & 2900.9 \\
            \midrule
            \multicolumn{7}{@{}l@{}}{\textbf{MODS (MDLM + DAPS)}} \\[1pt]
            $\alpha=2^8$ & 409.7 & 255.9 & 165.5 & 94.3 & 56.6 & 2431.5 \\
            $\alpha=2^{10}$ & 583.9 & 335.3 & 229.5 & 113.5 & 67.0 & 3602.0 \\
            $\alpha=2^{12}\,^*$ & 719.1 & 399.7 & 261.7 & {\cellcolor{bestcell}}\bfseries 129.9 & \bfseries 67.9 & 4660.0 \\
            $\alpha=2^{14}$ & {\cellcolor{bestcell}}\bfseries 791.6 & \bfseries 409.1 & {\cellcolor{bestcell}}\bfseries 287.8 & 117.1 & 62.3 & \cellcolorlast{bestcell}{\bfseries 4777.8} \\
            \midrule
            \multicolumn{7}{@{}l@{}}{\textbf{MODS (D3PM + CG)}} \\[1pt]
            $\tau=0.4^7$ & \bfseries 509.2 & 233.3 & \bfseries 127.7 & \bfseries 64.7 & \bfseries 58.8 & 1208.8 \\
            $\tau=0.4^6$ & 502.6 & \bfseries 246.3 & 121.8 & 57.2 & 56.4 & 1058.7 \\
            $\tau=0.4^5$ & 475.5 & 240.9 & 111.6 & 53.3 & 55.3 & 1083.3 \\
            $\tau=0.4^4$ & 384.7 & 236.7 & 116.9 & 50.1 & 52.7 & \bfseries 1259.0 \\
            \midrule
            \multicolumn{7}{@{}l@{}}{\textbf{MODS (D3PM + DAPS)}} \\[1pt]
            $\alpha=2^8$ & 574.4 & 299.9 & 157.1 & 67.3 & 65.9 & 1937.7 \\
            $\alpha=2^{10}$ & 676.3 & 380.2 & 223.1 & 90.0 & 80.7 & 3053.3 \\
            $\alpha=2^{12}$ & 754.6 & 415.5 & 256.5 & 104.3 & 85.4 & 4000.8 \\
            $\alpha=2^{14}\,^*$ & \bfseries 778.8 & {\cellcolor{bestcell}}\bfseries 420.1 & \bfseries 257.5 & \bfseries 107.0 & {\cellcolor{bestcell}}\bfseries 86.1 & \bfseries 4145.7 \\
            \bottomrule
            \multicolumn{7}{@{}l@{}}{
                \begin{minipage}{7cm}
                    \par\vspace{4pt}
                    \footnotesize
                    $\tau$: CG temperature; $\alpha$: DAPS steering strength. \\
                    $^*$: variants shown in Figure~\ref{fig:main_results}, Figure~\ref{fig:pareto-comparison} and Table~\ref{tab:results_hv}.
                \end{minipage}
            }
        \end{tabular}%
    \endgroup
\end{table}
\Cref{tab:hypervolume_hpo} shows the mean results for two restarts with separate random seeds, which were used to choose the steering strength for the main results.
For each configuration, we chose the steering parameter that resulted in the smallest maximum drop in hypervolume across tasks.

\begin{table}[p]
    \centering
    \caption{
    \textbf{Final hypervolume across all evaluated methods.} Entries report mean $\pm$ one standard error of the mean over five repetitions. Bold indicates the highest mean among different parameters of a method.
    Green highlights indicates the highest mean among all methods.
    }
    \label{tab:full_hypervolume_sweep}
    \begingroup
        \setlength{\tabcolsep}{4pt}
        \renewcommand{\arraystretch}{1.15}
        \setlastcellformat{M{4.1}{3.1}}
        \begin{tabular}{@{}l *3{M{3.1}{2.1}} M{3.1}{1.1} M{2.1}{1.1} M{4.1}{3.1} @{}}
            \toprule
            Setting & {GB1} & \shortstack{ERBIN\\PDZ1} & \shortstack{NHERF3\\PDZ2} & {KRAS} & {Src} & \shortstack{NHERF3\\PDZ1} \\

            \midrule
            \multicolumn{7}{@{}l}{\textbf{MDLM (unsteered)}} \\[1pt]
            & 231.6 \pm 11.4 & 231.4 \pm 2.0 & 145.4 \pm 5.0 & 70.0 \pm 6.1 & 64.0 \pm 2.4 & 2244.8 \pm 78.8 \\
            \midrule
            \multicolumn{7}{@{}l}{\textbf{D3PM (unsteered)}} \\[1pt]
            & 225.0 \pm 4.8 & 221.0 \pm 5.5 & 95.7 \pm 3.4 & 40.7 \pm 2.1 & 44.7 \pm 1.0 & 1004.0 \pm 28.4 \\
            \midrule
            \multicolumn{7}{@{}l}{\textbf{MODS (MDLM + CG)}} \\[1pt]
            $\tau=0.4^7$ & \bfseries 518.0 \pm 13.0 & \bfseries 318.4 \pm 14.6 & 184.2 \pm 11.8 & 99.0 \pm 2.5 & 78.1 \pm 4.7 & \bfseries 3873.3 \pm 92.0 \\
            $\tau=0.4^6$ & 458.9 \pm 7.7 & 316.5 \pm 8.8 & \bfseries 223.0 \pm 2.8 & \bfseries 110.4 \pm 1.6 & \bfseries 81.7 \pm 1.5 & 3718.9 \pm 30.8 \\
            $\tau=0.4^5$ & 405.5 \pm 9.5 & 295.8 \pm 7.0 & 210.2 \pm 2.3 & 108.8 \pm 1.3 & 76.8 \pm 5.0 & 3396.0 \pm 72.7 \\
            $\tau=0.4^4$ & 332.9 \pm 2.6 & 258.7 \pm 4.7 & 190.6 \pm 4.0 & 100.8 \pm 0.7 & 75.8 \pm 1.2 & 2865.6 \pm 32.9 \\
            \midrule
            \multicolumn{7}{@{}l}{\textbf{MODS (MDLM + DAPS)}} \\[1pt]
            $\alpha=2^8$ & 421.0 \pm 8.8 & 262.1 \pm 2.7 & 173.7 \pm 4.3 & 82.1 \pm 3.2 & 58.4 \pm 2.9 & 2380.5 \pm 41.8 \\
            $\alpha=2^{10}$ & 590.4 \pm 12.4 & 330.0 \pm 2.2 & 228.9 \pm 3.4 & 111.0 \pm 0.8 & 64.3 \pm 2.3 & 3566.7 \pm 65.2 \\
            $\alpha=2^{12}\,^*$ & 685.3 \pm 10.4 & 391.4 \pm 3.7 & 254.9 \pm 5.3 & {\cellcolor{bestcell}}\bfseries 128.2 \pm 3.6 & \bfseries 66.4 \pm 2.8 & 4403.5 \pm 97.0 \\
            $\alpha=2^{14}$ & {\cellcolor{bestcell}}\bfseries 801.6 \pm 15.3 & \bfseries 401.1 \pm 6.6 & \bfseries 262.1 \pm 6.1 & 110.7 \pm 3.0 & 61.4 \pm 0.6 & \cellcolorlast{bestcell}{\bfseries 4846.1 \pm 109.9} \\
            \midrule
            \multicolumn{7}{@{}l}{\textbf{MODS (D3PM + CG)}} \\[1pt]
            $\tau=0.4^7$ & \bfseries 490.7 \pm 7.5 & 241.3 \pm 4.4 & \bfseries 135.2 \pm 4.8 & \bfseries 66.1 \pm 1.7 & \bfseries 60.2 \pm 0.6 & \bfseries 1441.1 \pm 127.0 \\
            $\tau=0.4^6$ & 483.8 \pm 9.7 & 233.2 \pm 4.1 & 129.3 \pm 2.7 & 56.4 \pm 0.8 & 56.2 \pm 1.1 & 1187.5 \pm 75.9 \\
            $\tau=0.4^5$ & 434.3 \pm 8.0 & \bfseries 241.8 \pm 4.3 & 121.7 \pm 7.2 & 51.2 \pm 0.5 & 55.1 \pm 0.7 & 1177.6 \pm 28.6 \\
            $\tau=0.4^4$ & 370.5 \pm 5.4 & 241.5 \pm 4.5 & 114.2 \pm 2.2 & 47.7 \pm 0.7 & 53.5 \pm 1.2 & 1149.8 \pm 37.8 \\
            \midrule
            \multicolumn{7}{@{}l}{\textbf{MODS (D3PM + DAPS)}} \\[1pt]
            $\alpha=2^8$ & 564.6 \pm 3.5 & 296.0 \pm 7.5 & 160.5 \pm 3.7 & 62.6 \pm 1.2 & 64.7 \pm 1.6 & 1989.8 \pm 45.8 \\
            $\alpha=2^{10}$ & 647.3 \pm 5.9 & 381.0 \pm 3.3 & 218.0 \pm 2.3 & 92.8 \pm 1.9 & 79.9 \pm 0.7 & 3298.6 \pm 143.0 \\
            $\alpha=2^{12}$ & 743.3 \pm 5.7 & 406.5 \pm 5.4 & 259.4 \pm 4.8 & 100.7 \pm 1.3 & {\cellcolor{bestcell}}\bfseries 85.8 \pm 0.9 & 4159.2 \pm 60.1 \\
            $\alpha=2^{14}\,^*$ & \bfseries 781.4 \pm 5.9 & {\cellcolor{bestcell}}\bfseries 413.8 \pm 4.2 & {\cellcolor{bestcell}}\bfseries 271.5 \pm 4.1 & \bfseries 106.5 \pm 2.3 & 85.2 \pm 0.3 & \bfseries 4278.9 \pm 72.5 \\
            \midrule
            \multicolumn{7}{@{}l}{\textbf{ARLM + DPO}} \\[1pt]
            $\beta=0.02$ & 285.9 \pm 22.5 & 247.9 \pm 4.4 & 165.2 \pm 5.1 & 96.0 \pm 1.4 & 65.8 \pm 1.9 & 2205.4 \pm 143.4 \\
            $\beta=0.1$ & \bfseries 328.4 \pm 6.8 & \bfseries 252.4 \pm 5.3 & \bfseries 165.7 \pm 2.1 & 97.5 \pm 1.6 & \bfseries 67.5 \pm 2.4 & \bfseries 2272.7 \pm 39.7 \\
            $\beta=0.5$ & 281.5 \pm 7.5 & 239.9 \pm 2.4 & 165.3 \pm 3.9 & \bfseries 98.5 \pm 2.5 & 65.1 \pm 2.1 & 2214.5 \pm 121.6 \\
            $\beta=2$ & 245.6 \pm 6.2 & 241.2 \pm 2.1 & 160.4 \pm 3.8 & 77.2 \pm 6.0 & 62.9 \pm 1.5 & 1859.1 \pm 70.3 \\
            \midrule
            \multicolumn{7}{@{}l}{\textbf{SAGE-Prot}} \\[1pt]
            {} & 210.8 \pm 2.5 & 230.5 \pm 2.6 & 130.2 \pm 9.1 & 95.6 \pm 1.3 & 66.2 \pm 3.2 & 1960.6 \pm 126.3 \\
            \midrule
            \multicolumn{7}{@{}l}{\textbf{LaMBO-2}} \\[1pt]
            {} & 234.2 \pm 13.5 & 216.2 \pm 12.8 & 121.4 \pm 6.4 & 44.1 \pm 2.2 & 45.7 \pm 1.2 & 1296.8 \pm 136.3 \\
            \bottomrule
            \multicolumn{7}{@{}l@{}}{
                \begin{minipage}{10cm}
                    \par\vspace{4pt}
                    \footnotesize
                    $\tau$: CG temperature; $\alpha$: DAPS steering strength;
                    $\beta$: DPO coefficient.
                    \\
                    $^*$: variants shown in Figure~\ref{fig:main_results}, Figure~\ref{fig:pareto-comparison} and Table~\ref{tab:results_hv}.
                \end{minipage}
            }
        \end{tabular}%
    \endgroup
\end{table}

\begin{table}
    \centering
    \caption{
        \textbf{$p$-values against baselines.} Entries compare the MODS configurations in Table~\ref{tab:results_hv} against each baseline for each problem.
        All $p$-values, except for the results for MODS (MDLM + DAPS) on the Src problem, are below $0.05$.
    }
    \label{tab:main_hypervolume_pvalues}
    \begingroup
        \setlength{\tabcolsep}{3pt} 
        \renewcommand{\arraystretch}{1.15}
        \sisetup{
          scientific-notation = true,
          exponent-product    = \cdot, 
          tight-spacing       = true,
        }

\begin{tabular}{@{} l l S[table-format=1.2e-2] *5{S[table-format=1.2e-1]} @{}}
\toprule
\shortstack{Gener.\\model} & Baseline & {GB1} & \shortstack{ERBIN\\PDZ1} & \shortstack{NHERF3\\PDZ2} & {KRAS} & {Src} & \shortstack{NHERF3\\PDZ1} \\
\midrule
\multirow{4}{*}{\rotatebox{90}{MDLM}}
& MDLM (unsteered) & 2.20e-09 & 1.54e-08 & 3.93e-07 & 1.16e-04 & 5.34e-01 & 2.02e-07 \\
& ARLM + DPO & 1.97e-08 & 9.22e-08 & 1.37e-05 & 2.38e-04 & 7.73e-01 & 3.09e-06 \\
& SAGE-Prot & 4.50e-07 & 2.46e-09 & 1.32e-05 & 3.58e-04 & 9.64e-01 & 6.19e-07 \\
& LaMBO-2 & 1.09e-08 & 7.29e-05 & 3.23e-07 & 3.69e-07 & 7.59e-04 & 2.31e-07 \\
\midrule
\multirow{4}{*}{\rotatebox[origin=c]{90}{D3PM}}
& D3PM (unsteered) & 3.33e-12 & 7.86e-09 & 1.33e-09 & 2.94e-08 & 4.34e-07 & 8.68e-08 \\
& ARLM + DPO & 3.95e-11 & 2.00e-08 & 4.77e-07 & 4.65e-02 & 1.64e-03 & 2.21e-07 \\
& SAGE-Prot & 9.47e-10 & 5.55e-09 & 1.42e-05 & 5.52e-03 & 3.88e-03 & 2.25e-06 \\
& LaMBO-2 & 8.63e-08 & 3.33e-05 & 2.89e-07 & 4.88e-08 & 1.78e-06 & 1.06e-06 \\
\bottomrule
\end{tabular}
    \endgroup
\end{table}

\Cref{tab:full_hypervolume_sweep} summarizes all MODS and baseline results over five restarts.
\Cref{tab:main_hypervolume_pvalues} shows related $p$-values that compare MODS (MDLM + DAPS) and MODS (D3PM + DAPS) against each baseline.

\begin{figure*}[t]
    \centering
    \begin{subfigure}[t]{0.48\linewidth}
        \centering
        \includegraphics[width=\linewidth]{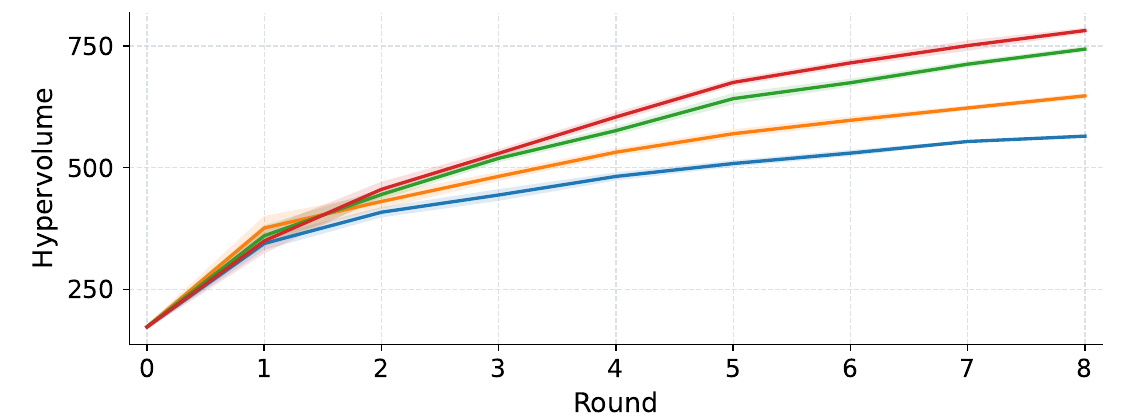}
        \caption{GB1, MODS (D3PM + DAPS)}
    \end{subfigure}
    \hfill
    \begin{subfigure}[t]{0.48\linewidth}
        \centering
        \includegraphics[width=\linewidth]{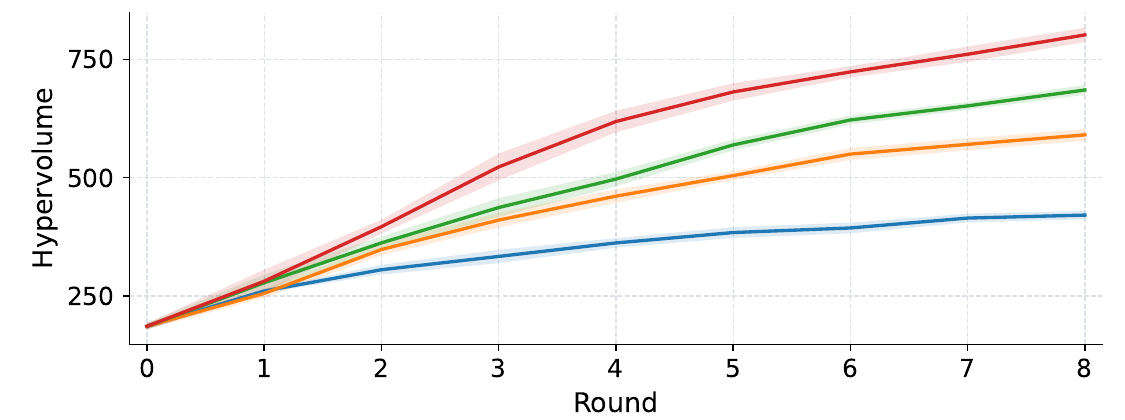}
        \caption{GB1, MODS (MDLM + DAPS)}
    \end{subfigure}
    \\[1ex]
    \begin{subfigure}[t]{0.48\linewidth}
        \centering
        \includegraphics[width=\linewidth]{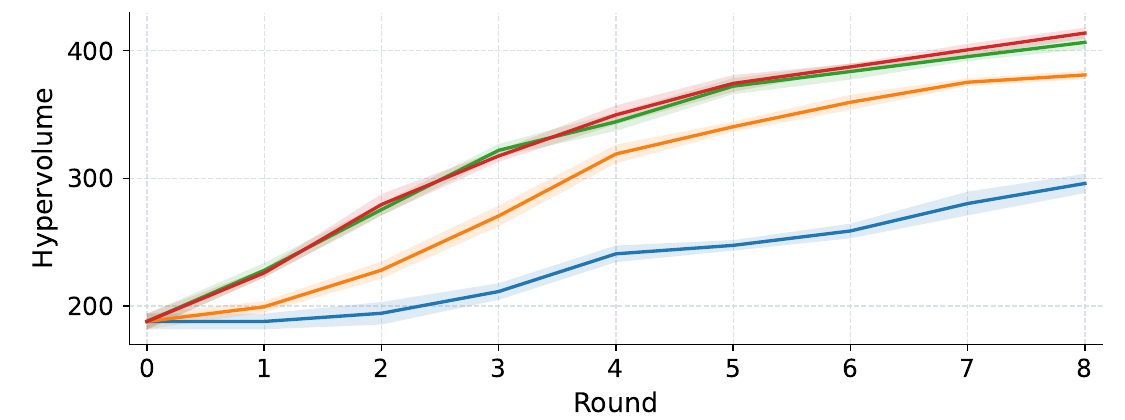}
        \caption{ERBIN-PDZ1, MODS (D3PM + DAPS)}
    \end{subfigure}
    \hfill
    \begin{subfigure}[t]{0.48\linewidth}
        \centering
        \includegraphics[width=\linewidth]{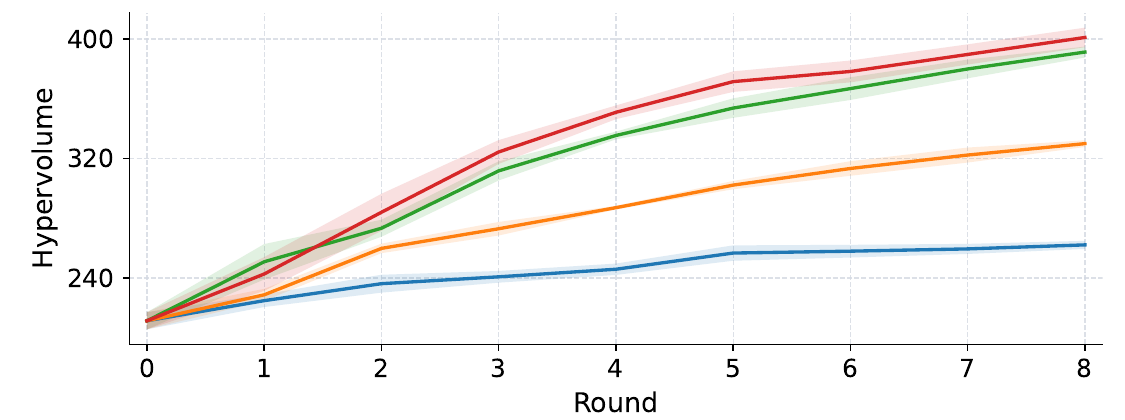}
        \caption{ERBIN-PDZ1, MODS (MDLM + DAPS)}
    \end{subfigure}
    \\[1ex]
    \begin{subfigure}[t]{0.48\linewidth}
        \centering
        \includegraphics[width=\linewidth]{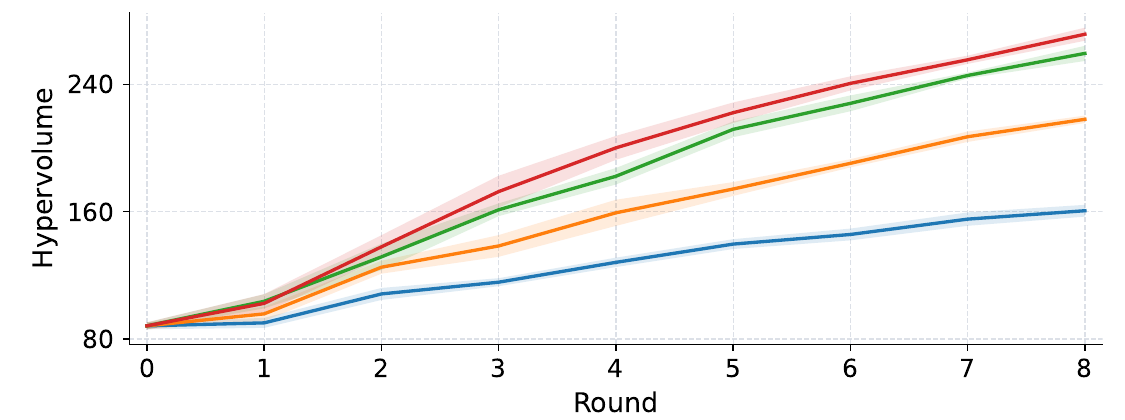}
        \caption{NHERF3-PDZ2, MODS (D3PM + DAPS)}
    \end{subfigure}
    \hfill
    \begin{subfigure}[t]{0.48\linewidth}
        \centering
        \includegraphics[width=\linewidth]{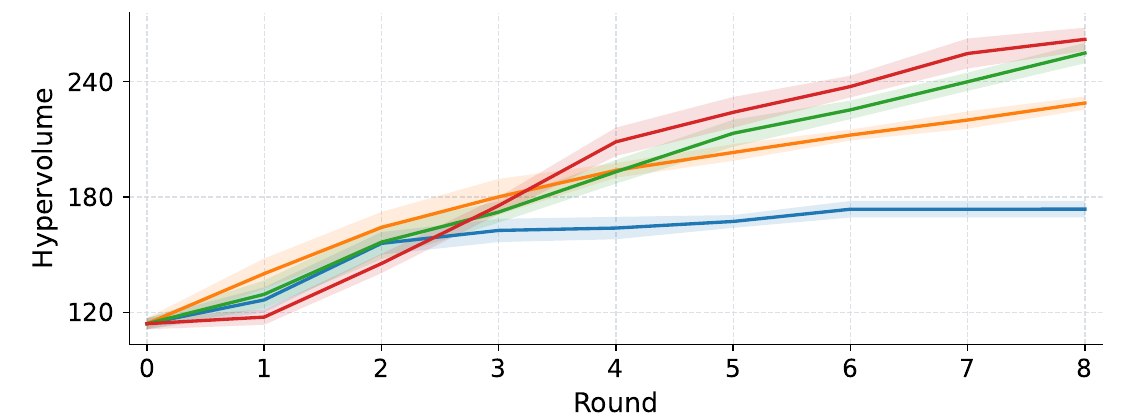}
        \caption{NHERF3-PDZ2, MODS (MDLM + DAPS)}
    \end{subfigure}
    \\[1ex]
    \begin{subfigure}[t]{0.48\linewidth}
        \centering
        \includegraphics[width=\linewidth]{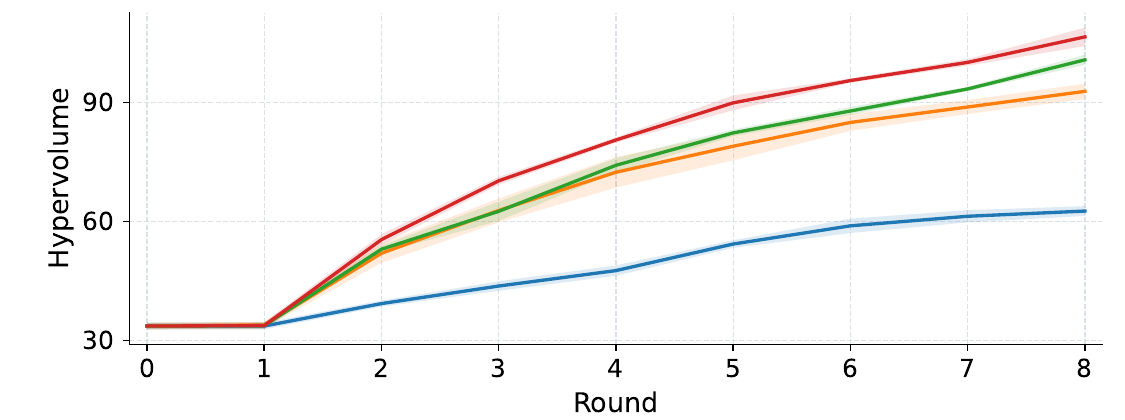}
        \caption{KRAS, MODS (D3PM + DAPS)}
    \end{subfigure}
    \hfill
    \begin{subfigure}[t]{0.48\linewidth}
        \centering
        \includegraphics[width=\linewidth]{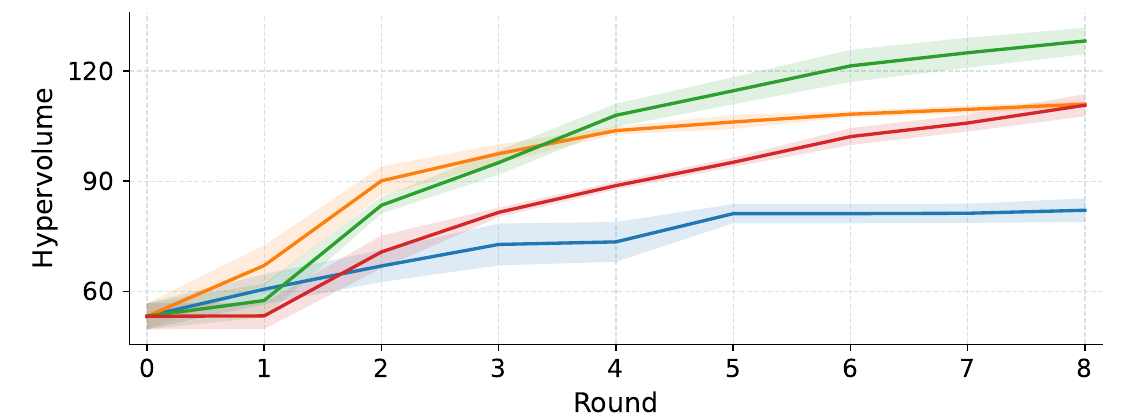}
        \caption{KRAS, MODS (MDLM + DAPS)}
    \end{subfigure}
    \\[1ex]
    \begin{subfigure}[t]{0.48\linewidth}
        \centering
        \includegraphics[width=\linewidth]{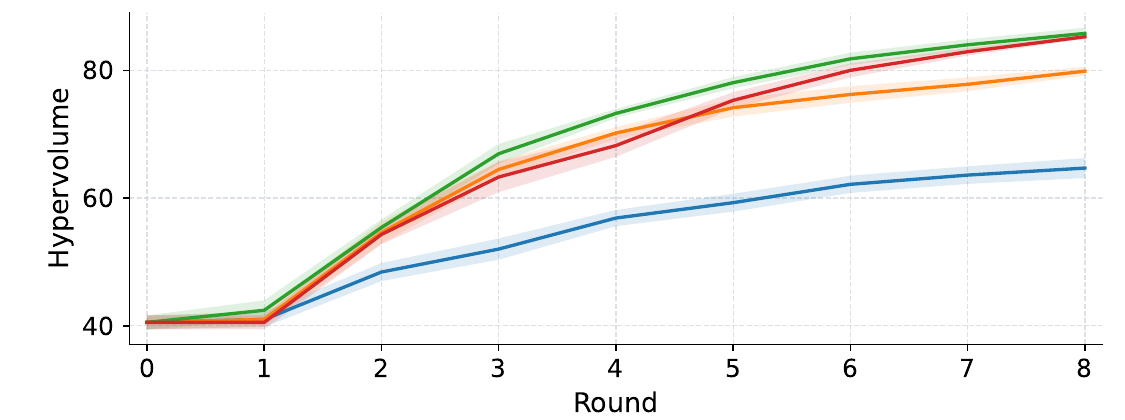}
        \caption{Src, MODS (D3PM + DAPS)}
    \end{subfigure}
    \hfill
    \begin{subfigure}[t]{0.48\linewidth}
        \centering
        \includegraphics[width=\linewidth]{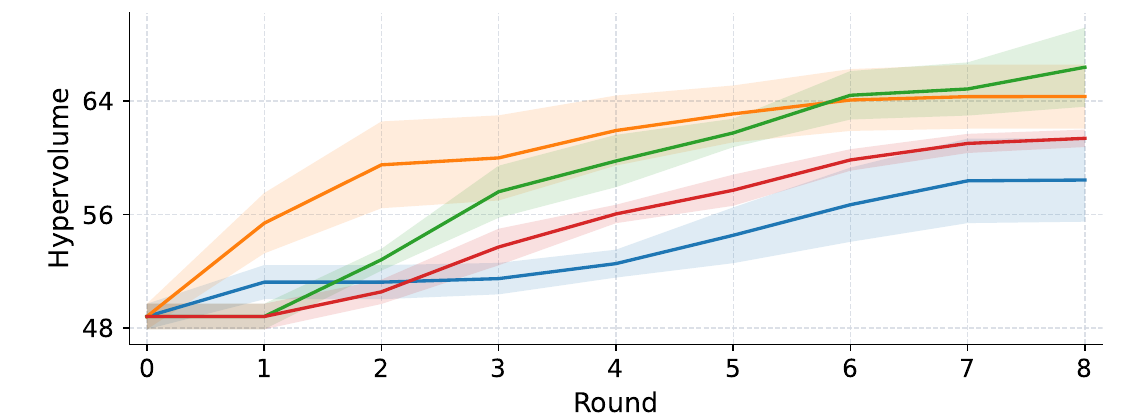}
        \caption{Src, MODS (MDLM + DAPS)}
    \end{subfigure}
    \\[1ex]
    \begin{subfigure}[t]{0.48\linewidth}
        \centering
        \includegraphics[width=\linewidth]{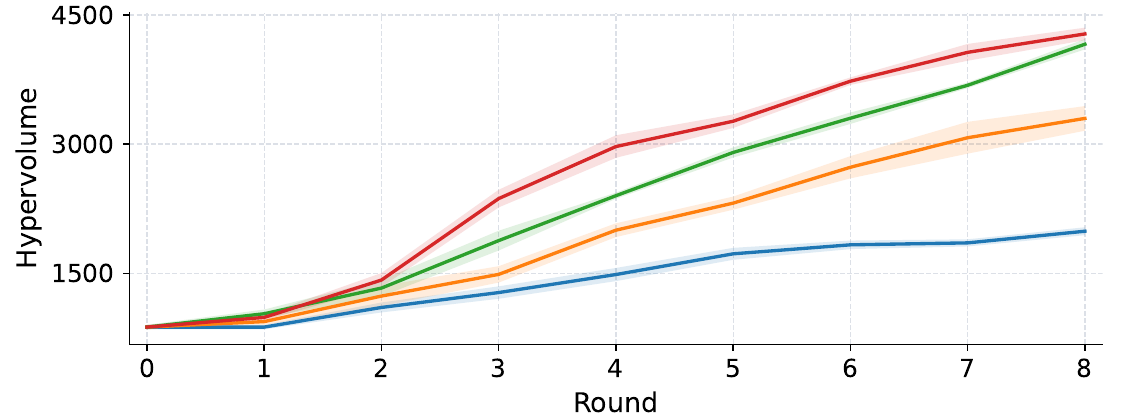}
        \caption{NHERF3-PDZ1, MODS (D3PM + DAPS)}
    \end{subfigure}
    \hfill
    \begin{subfigure}[t]{0.48\linewidth}
        \centering
        \includegraphics[width=\linewidth]{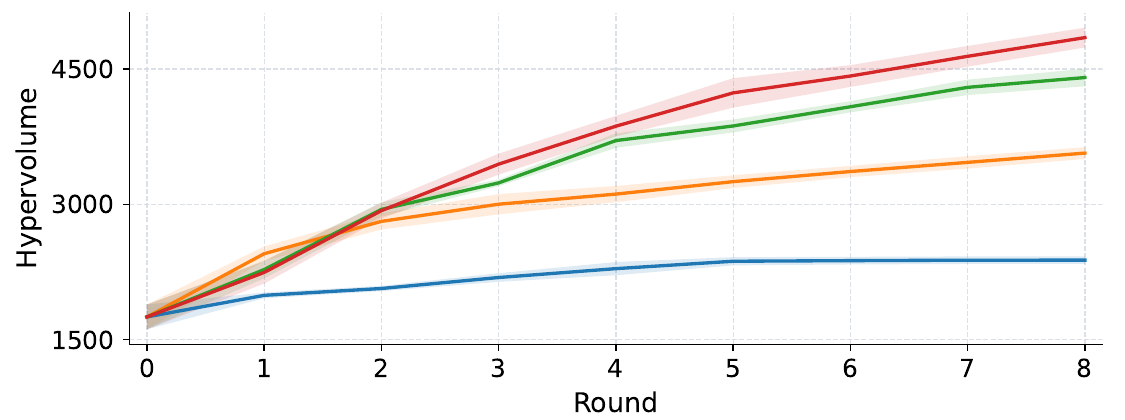}
        \caption{NHERF3-PDZ1, MODS (MDLM + DAPS)}
    \end{subfigure}
    \centering
    \includegraphics[width=0.75\textwidth]{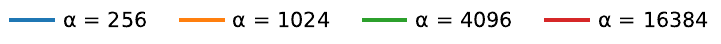}
    \caption{
        \textbf{Sensitivity of DAPS to the steering strength.} Large $\alpha$ values correspond to weak regularization and strong steering. Lines indicate the mean hypervolume over five repeats. Shaded regions denote the mean $\pm$ one standard error of the mean over five repetitions.
    }
    \label{fig:ablation-daps}
\end{figure*}

\begin{figure*}[t]
    \centering
    \begin{subfigure}[t]{0.48\linewidth}
        \centering
        \includegraphics[width=\linewidth]{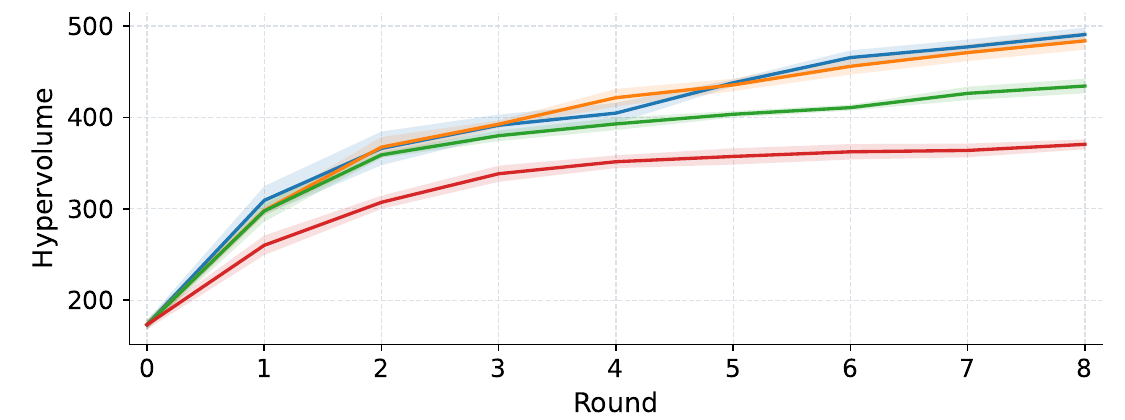}
        \caption{GB1, MODS (D3PM + CG)}
    \end{subfigure}
    \hfill
    \begin{subfigure}[t]{0.48\linewidth}
        \centering
        \includegraphics[width=\linewidth]{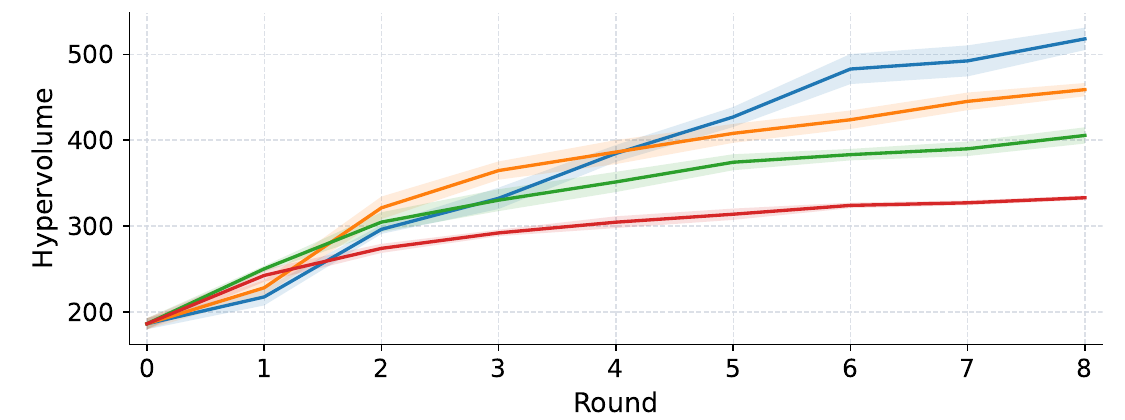}
        \caption{GB1, MODS (MDLM + CG)}
    \end{subfigure}
    \\[1ex]
    \begin{subfigure}[t]{0.48\linewidth}
        \centering
        \includegraphics[width=\linewidth]{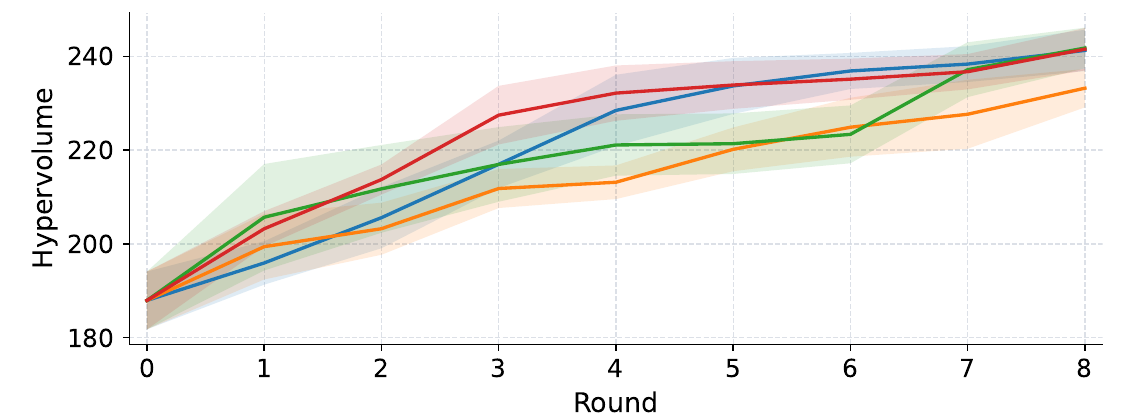}
        \caption{ERBIN-PDZ1, MODS (D3PM + CG)}
    \end{subfigure}
    \hfill
    \begin{subfigure}[t]{0.48\linewidth}
        \centering
        \includegraphics[width=\linewidth]{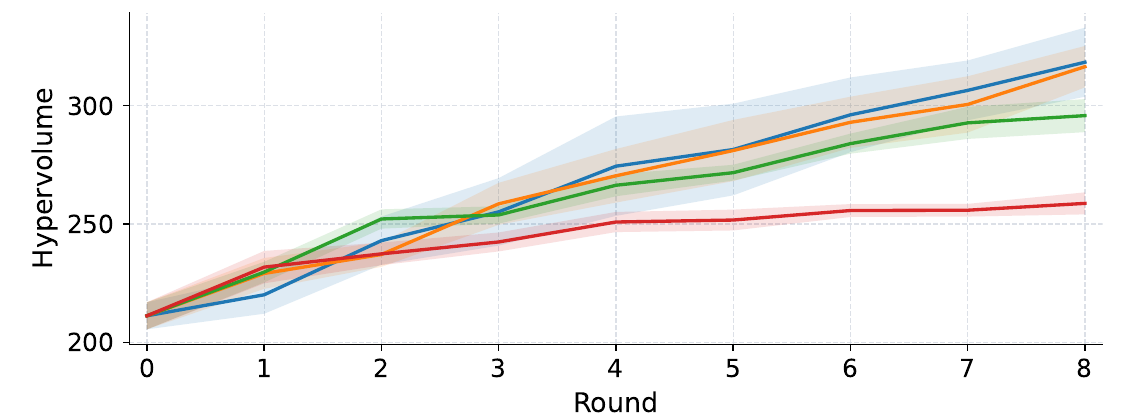}
        \caption{ERBIN-PDZ1, MODS (MDLM + CG)}
    \end{subfigure}
    \\[1ex]
    \begin{subfigure}[t]{0.48\linewidth}
        \centering
        \includegraphics[width=\linewidth]{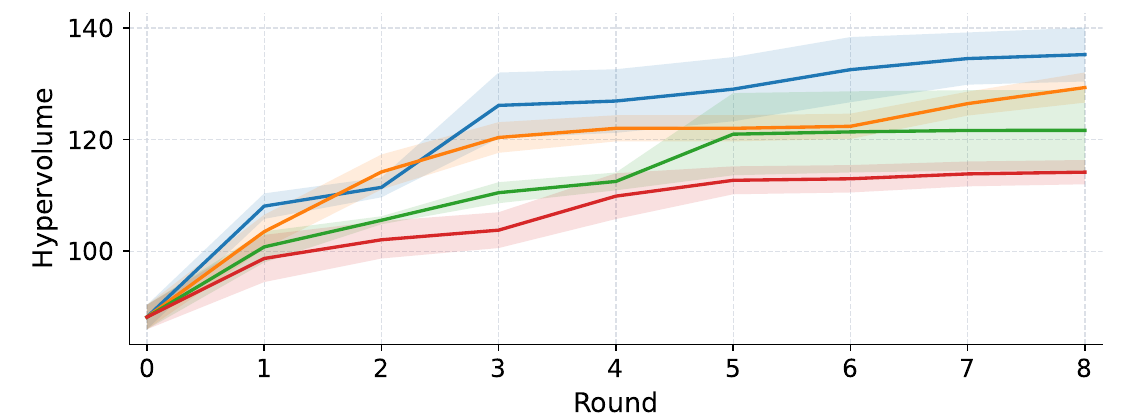}
        \caption{NHERF3-PDZ2, MODS (D3PM + CG)}
    \end{subfigure}
    \hfill
    \begin{subfigure}[t]{0.48\linewidth}
        \centering
        \includegraphics[width=\linewidth]{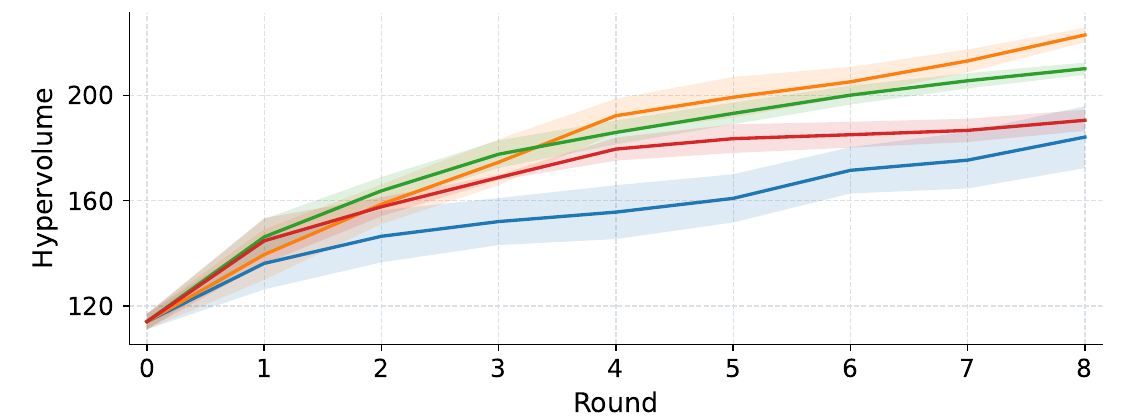}
        \caption{NHERF3-PDZ2, MODS (MDLM + CG)}
    \end{subfigure}
    \\[1ex]
    \begin{subfigure}[t]{0.48\linewidth}
        \centering
        \includegraphics[width=\linewidth]{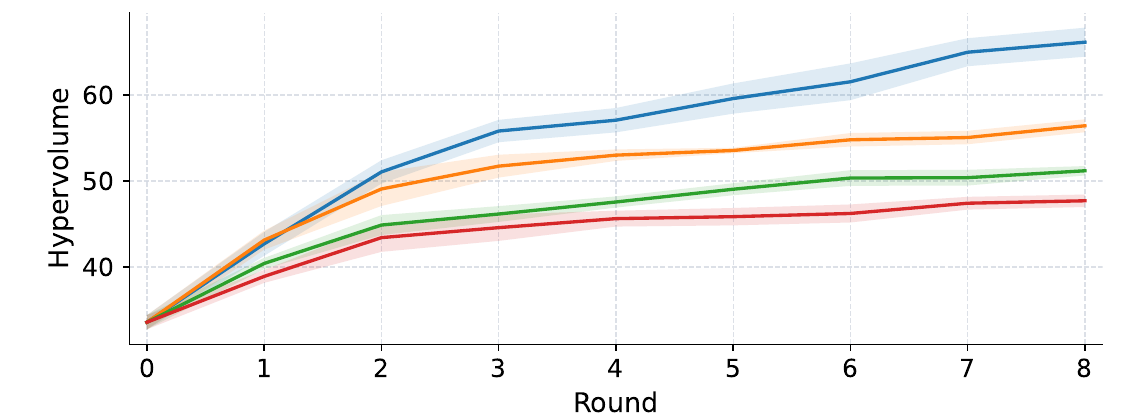}
        \caption{KRAS, MODS (D3PM + CG)}
    \end{subfigure}
    \hfill
    \begin{subfigure}[t]{0.48\linewidth}
        \centering
        \includegraphics[width=\linewidth]{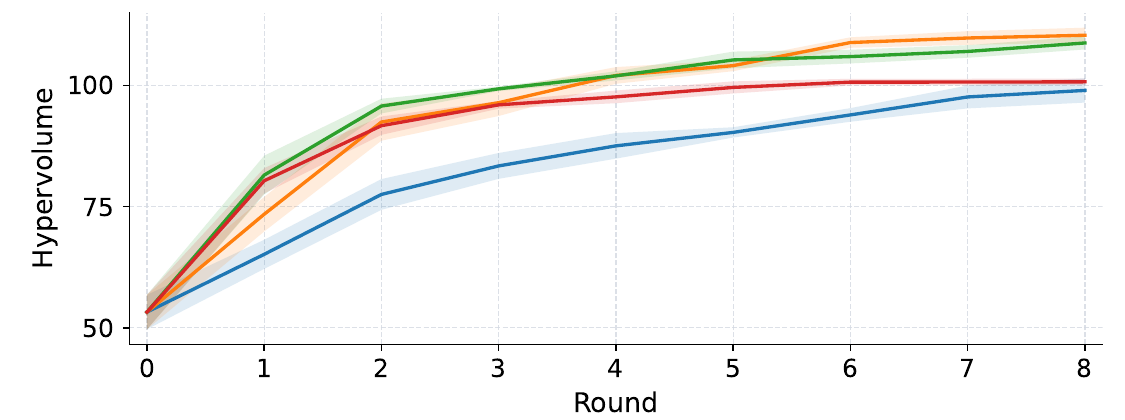}
        \caption{KRAS, MODS (MDLM + CG)}
    \end{subfigure}
    \\[1ex]
    \begin{subfigure}[t]{0.48\linewidth}
        \centering
        \includegraphics[width=\linewidth]{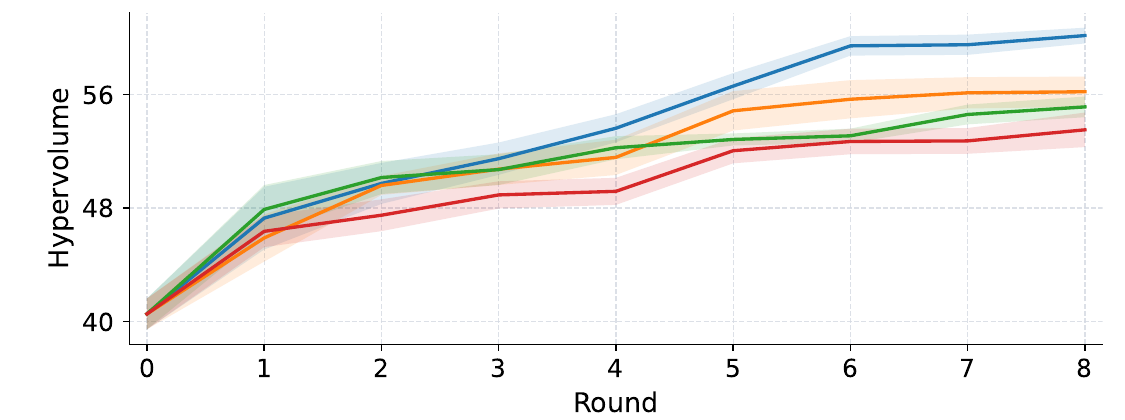}
        \caption{Src, MODS (D3PM + CG)}
    \end{subfigure}
    \hfill
    \begin{subfigure}[t]{0.48\linewidth}
        \centering
        \includegraphics[width=\linewidth]{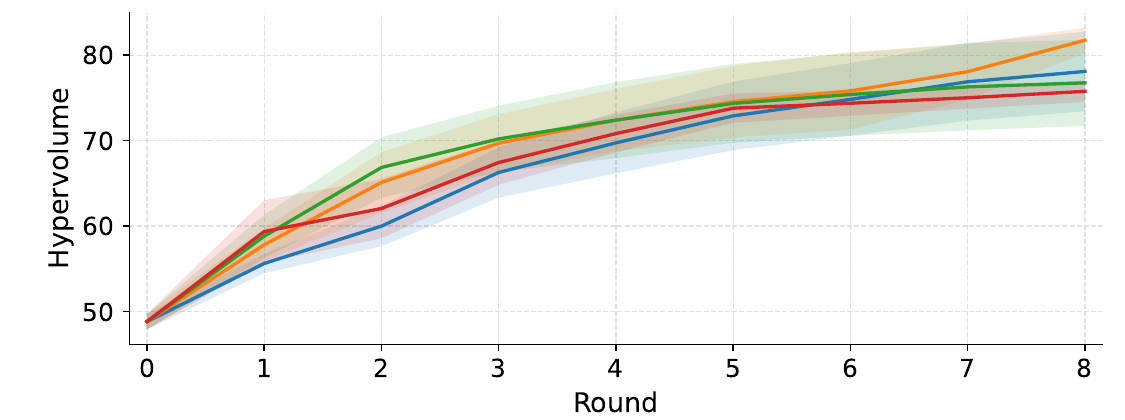}
        \caption{Src, MODS (MDLM + CG)}
    \end{subfigure}
    \\[1ex]
    \begin{subfigure}[t]{0.48\linewidth}
        \centering
        \includegraphics[width=\linewidth]{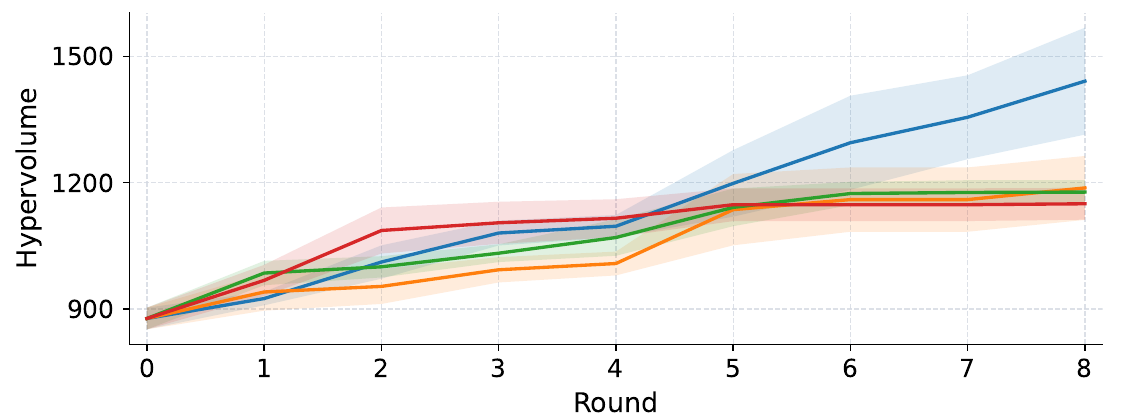}
        \caption{NHERF3-PDZ1, MODS (D3PM + CG)}
    \end{subfigure}
    \hfill
    \begin{subfigure}[t]{0.48\linewidth}
        \centering
        \includegraphics[width=\linewidth]{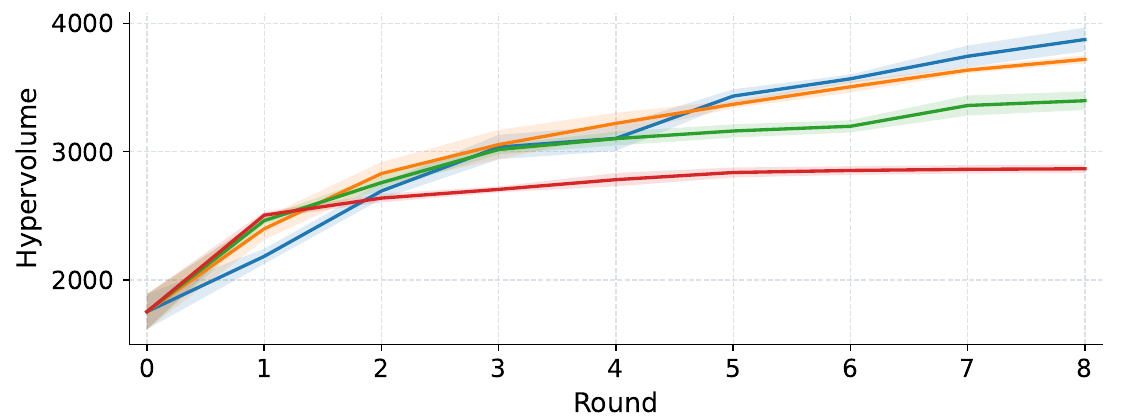}
        \caption{NHERF3-PDZ1, MODS (MDLM + CG)}
    \end{subfigure}
    \centering
    \includegraphics[width=0.75\textwidth]{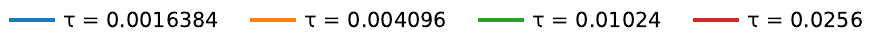}
    \caption{
        \textbf{Sensitivity of classifier guidance to the steering strength.} Small $\tau \coloneqq \alpha^{-1}$ values correspond to weak regularization and strong steering. Lines indicate the mean hypervolume over five repeats. Shaded regions denote the mean $\pm$ one standard error of the mean over five repetitions.
    }
    \label{fig:ablation-cg}
\end{figure*}

Figures~\ref{fig:ablation-daps} and~\ref{fig:ablation-cg} show the effect of different steering strengths on the hypervolume over multiple rounds.

\section{Extended optimization results}
\label{sec:additional_results}

\begin{table}[t]
    \centering
    \caption{
    \textbf{Computational cost of the steering methods.} Entries report mean $\pm$ one standard deviation in seconds. Generation cost is independent of the round, whereas surrogate training grows with the size of the accumulated labelled pool. All experiments were run on a single NVIDIA RTX A6000 (48\,GB).
    }
    \label{tab:computational_cost}
    \begingroup
        \setlength{\tabcolsep}{4pt} 
        \begin{tabular}{@{}l M{4.0}{3.0} M{3.0}{2.0} M{4.0}{3.0} M{3.0}{2.0} @{}}
            \toprule
            & \multicolumn{1}{c}{MDLM + CG} & \multicolumn{1}{c}{MDLM + DAPS} & \multicolumn{1}{c}{D3PM + CG} & \multicolumn{1}{c@{}}{D3PM + DAPS} \\
            \midrule
            \multicolumn{5}{@{}l@{}}{\textbf{Round 1}} \\[1pt]
            \quad Surrogate training [\unit{\second}] & 151 \pm 21 & 29 \pm 6 & 139 \pm 17 & 28 \pm 6 \\
            \quad Generation [\unit{\second}] & 49 \pm 28 & 617 \pm 30 & 152 \pm 109 & 600 \pm 40 \\
            \quad Total [\unit{\second}] & 200 \pm 38 & 646 \pm 31 & 291 \pm 112 & 628 \pm 42 \\
            \midrule
            \multicolumn{5}{@{}l@{}}{\textbf{Round 8}} \\[1pt]
            \quad Surrogate training [\unit{\second}] & 955 \pm 130 & 191 \pm 34 & 899 \pm 69 & 180 \pm 33 \\
            \quad Generation [\unit{\second}] & 49 \pm 28 & 617 \pm 30 & 152 \pm 109 & 600 \pm 40 \\
            \quad Total [\unit{\second}] & 1004 \pm 144 & 808 \pm 48 & 1051 \pm 115 & 780 \pm 65 \\
            \bottomrule
        \end{tabular}
    \endgroup
\end{table}
\Cref{tab:computational_cost} shows the run time of different MODS configurations.
The run time of training surrogates increases over optimization rounds as more data is acquired and used for training.
Surrogate training for CG is generally more expensive because the surrogates are conditioned on the denoising step (see Appendix~\ref{sec:surrogate_details}).
Sample generation with DAPS is more expensive than CG due to the Metropolis-Hastings steps in each denoising step.

\begin{figure}[t]
    \begin{center}
        \includegraphics[width=0.99\linewidth]{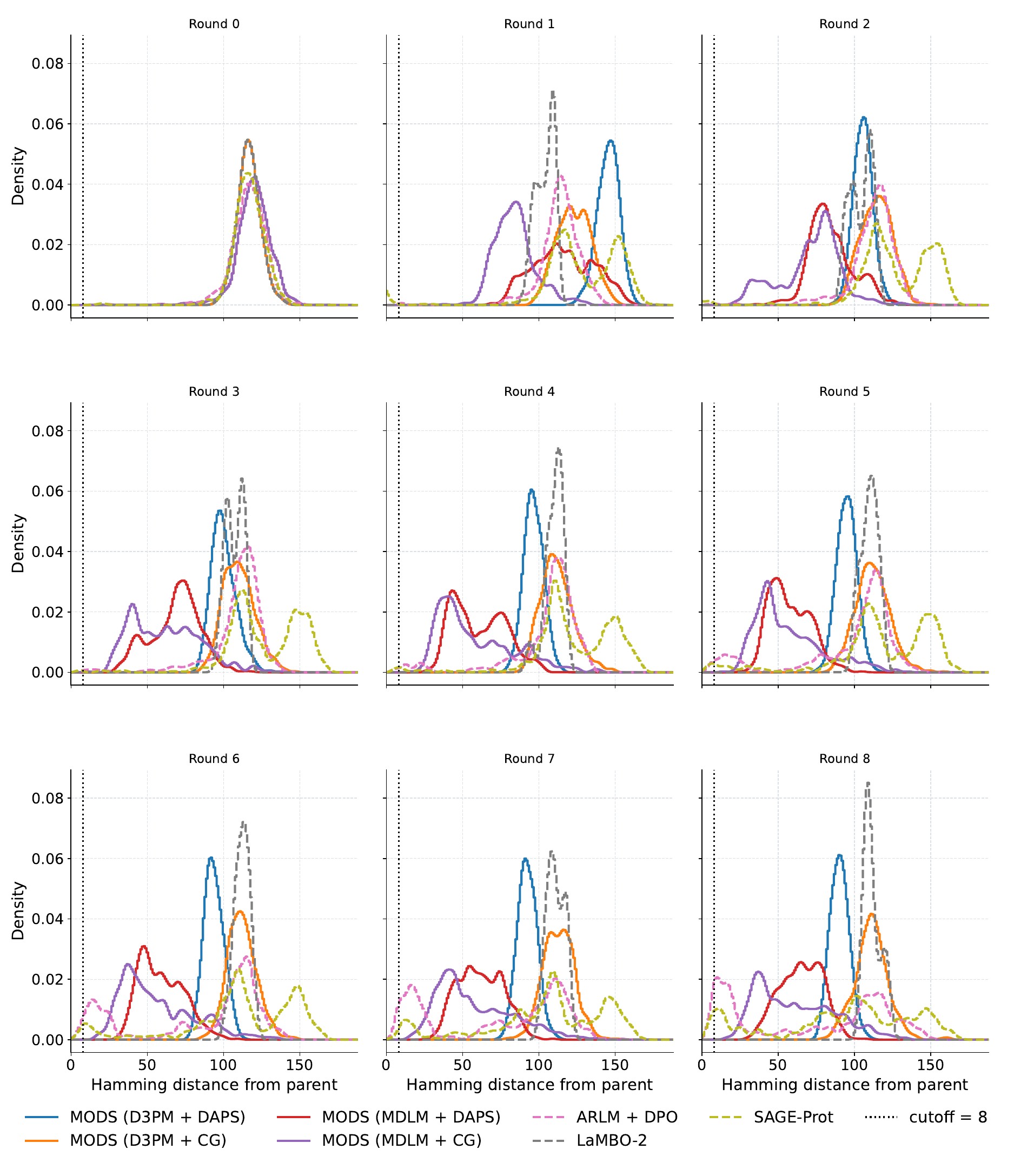}
    \end{center}
    \caption{\textbf{Hamming distance evolution of generated proteins throughout the campaign.} Each panel shows the distribution of distances from the parent sequence for the evaluated sequences in each round.
    The dotted vertical line marks the distance threshold $d_0=8$.}
    \label{fig:kras_hd}
\end{figure}

Figure~\ref{fig:kras_hd} shows how the Hamming distance of sampled sequences from the parent evolves across the 8 rounds for the KRAS task. The dotted vertical line marks the threshold \(d_0 = 8\) above which the distance penalty from Appendix~\ref{sec:fitness_details} is applied.

The distance distributions show that different methods, generative models, and steering mechanisms yield different evaluations in the sequence space.

For all methods, a substantial part of the evaluated sequences are beyond the $d_0$ distance for the penalty. The penalty, therefore, acts as a soft penalty to sequences far away from the base sequence, allowing evaluations outside the mutational neighborhood of the oracle training data.

For MODS with MDLM, the generative model yields a distribution of Hamming distances that is closer to the base sequence but generally wider.
For MODS with D3PM the distances are generally larger but more concentrated.

Comparing these results with the achieved hypervolumes in Figure~\ref{fig:main_results}, shows no direct relationship between the distance of the parent sequence and the hypervolume.


\end{document}